\documentclass[Afour,sageh,times]{sagej}

\usepackage{moreverb,url}

\usepackage{times}
\usepackage{multicol}
\usepackage{comment}
\usepackage{siunitx}
\usepackage{relsize}
\usepackage{ifthen}
\usepackage[colorinlistoftodos]{todonotes}
\usepackage[vlined,ruled,linesnumbered]{algorithm2e}
\usepackage{graphics} 
\usepackage{rotating}
\usepackage{color}
\usepackage{enumerate}
\usepackage[T1]{fontenc}
\usepackage{psfrag}
\usepackage{epsfig} 
\usepackage{booktabs}
\usepackage{graphicx}
\usepackage{url}
\usepackage{multirow}
\usepackage{array}
\usepackage{latexsym}
\usepackage{amsfonts}
\usepackage{amsmath}
\usepackage{amssymb}
\usepackage{dsfont}
\usepackage{xstring}
\usepackage{algorithmic}
\usepackage{multirow}
\usepackage{xcolor}
\usepackage{prettyref}
\usepackage{flexisym}
\usepackage{bigdelim}
\usepackage{breqn} 
\usepackage{listings}

\usepackage{xspace}
\usepackage{bm}
\usepackage{soul}
\graphicspath{{./figures/}}
\usepackage{tikz}
\usetikzlibrary{matrix,calc}
\usepackage{lipsum}
\usepackage{mdwlist}

\makecompactlist{itemize}{stditemize}
\usepackage{enumitem}
\usepackage{caption}
\usepackage{epstopdf}
\usepackage{subfigure}

\usepackage{amsthm}
\usepackage{mathtools}

\usepackage[
    colorlinks,
    bookmarksopen,
    bookmarksnumbered,
    citecolor=red,
    urlcolor=red
]{hyperref}

\usepackage{cleveref}

\usepackage[colorlinks,bookmarksopen,bookmarksnumbered,citecolor=red,urlcolor=red]{hyperref}

\newcommand\BibTeX{{\rmfamily B\kern-.05em \textsc{i\kern-.025em b}\kern-.08em
T\kern-.1667em\lower.7ex\hbox{E}\kern-.125emX}}

\def\volumeyear{2026}
\newtheorem{theorem}{Theorem}
\newtheorem{problem}{Problem}

\newtheorem{corollary}{Corollary}

\newtheorem{lemma}{Lemma}
\newtheorem{assumption}{Assumption}
\newtheorem{definition}{Definition}
\newtheorem{proposition}{Proposition}

\newtheorem{remark}{Remark}

\newcommand{\bdmath}{\begin{dmath}}
\newcommand{\edmath}{\end{dmath}}
\newcommand{\beq}{\begin{equation}}
\newcommand{\eeq}{\end{equation}}
\newcommand{\bdm}{\begin{displaymath}}
\newcommand{\edm}{\end{displaymath}}
\newcommand{\bea}{\begin{eqnarray}}
\newcommand{\eea}{\end{eqnarray}}
\newcommand{\beal}{\beq \begin{array}{lll}}
\newcommand{\eeal}{\end{array} \eeq}
\newcommand{\beas}{\begin{eqnarray*}}
\newcommand{\eeas}{\end{eqnarray*}}
\newcommand{\ba}{\begin{array}}
\newcommand{\ea}{\end{array}}
\newcommand{\bit}{\begin{itemize}}
\newcommand{\eit}{\end{itemize}}
\newcommand{\ben}{\begin{enumerate}}
\newcommand{\een}{\end{enumerate}}

\newcommand{\calD}{{\cal D}}

\newcommand{\calF}{{\cal F}}

\newcommand{\calH}{{\cal H}}

\newcommand{\calO}{{\cal O}}

\newcommand{\calU}{{\cal U}}

\definecolor{myblue}{RGB}{65 105 225}

\newcommand{\hide}[1]{}

\newcommand{\hiddenText}{{\color{gray} hidden text.}}
\newcommand{\hideWithText}[1]{\hiddenText}

\newcommand{\scenario}[1]{{\fontsize{9}{8.7}\selectfont\sf#1}\xspace}

\newcommand{\ie}{\emph{i.e.},\xspace}
\newcommand{\eg}{\emph{e.g.},\xspace}

\newcommand{\myParagraph}[1]{{\it #1:}\xspace}

\newcommand{\RKHS}{\scenario{{RKHS}}}
\newcommand{\OGD}{\scenario{{OGD}}}

\newcommand{\MPC}{\scenario{{MPC}}}

\newcommand{\DReg}{\operatorname{Regret}_T^D}

\newcommand{\RegOGD}{\operatorname{Regret}_{OGD}}
\newcommand{\SSI}{\scenario{{RFF-MPC}}}
\newcommand{\Hedge}{\scenario{{Hedge}}}
\newcommand{\RFF}{\scenario{{RFF}}}
\newcommand{\ISO}{\scenario{{ISO}}}
\newcommand{\AS}{\scenario{{AS}}}
\newcommand{\predA}{\scenario{{P1}}}
\newcommand{\predB}{\scenario{{P2}}}
\newcommand{\predC}{\scenario{{P3}}}
\newcommand{\predD}{\scenario{{P4}}}
\newcommand{\OPT}{\scenario{{Optimal*}}}
\newcommand{\MODL}{\scenario{{MODL}}}
\newcommand{\PLOT}{\scenario{{PLOT}}}
\newcommand{\FSF}{\scenario{{FSF}}}

\begin{document}

\runninghead{Navsalkar*, Zhou*, Tzoumas}

\title{Self-Adaptive Learning and Model Predictive Control for Tracking Unknown Dynamics with No Regret}

\author{Atharva Navsalkar*, Hongyu Zhou*, Vasileios Tzoumas \\ 
}

\affiliation{* Equal Contribution \\ 
The authors are with the Department of Aerospace Engineering, University of Michigan, Ann Arbor, MI 48109 USA \\
This work was supported in part by the National Science Foundation (NSF) CAREER Award No. 2337412, and the Army Research Office (ARO) Early Career Program Award W911NF-25-1-0280
}

\corrauth{Hongyu Zhou,
Department of Aerospace Engineering, 
University of Michigan, 
Ann Arbor, MI 48109 USA.}

\email{zhouhy@umich.edu}

\begin{abstract}
We propose a self-adaptive online learning for control method for tracking unknown target dynamics, where the target dynamics can exhibit switching behavior. 
Such challenging target tracking scenarios arise in applications of dynamic mapping, traffic control, and pursuit evasion, where robots need to track, pursue, or avoid collision with moving landmarks, objects, humans, etc., whose dynamics are unknown. 
Our method simultaneously learns multiple predictors from scratch, via self-supervised, one-shot, and computationally efficient learning, and adaptively selects the best one to match the observed target behavior. 
We enable our method via (i) random Fourier features (\RFF) for approximating the unknown target dynamics of each prediction model, (ii) online convex optimization for least-squares learning of the \RFF parameters, and (iii) bandit learning with delayed feedback to automatically select the best learned predictor. 
The method enjoys finite-time near-optimality guarantees in expectation, characterized as a function of the learning error of the target dynamics and the frequency that the target dynamics switch.  In the absence of both error and switching, the method asymptotically matches the optimal non-causal control policy that knows \textit{a priori} the target dynamics, \ie the method enjoys no regret in expectation. 
In the presence of learning errors and switching, the method degrades gracefully, \eg when there are errors and no switching, the average regret is proportional to the learning error.
We validate our method in Crazyflie simulations and hardware experiments, across target trajectories that vary from structured to random to adversarial, in comparison to non-stochastic, kernel-based, and neural-network-based methods for online learning. 
Averaged across all the targets, any single baseline has at least $38\%$ higher average tracking error than our method.
\end{abstract}


\maketitle

\begin{figure*}[t]
    \centering

    \includegraphics[width=\textwidth]{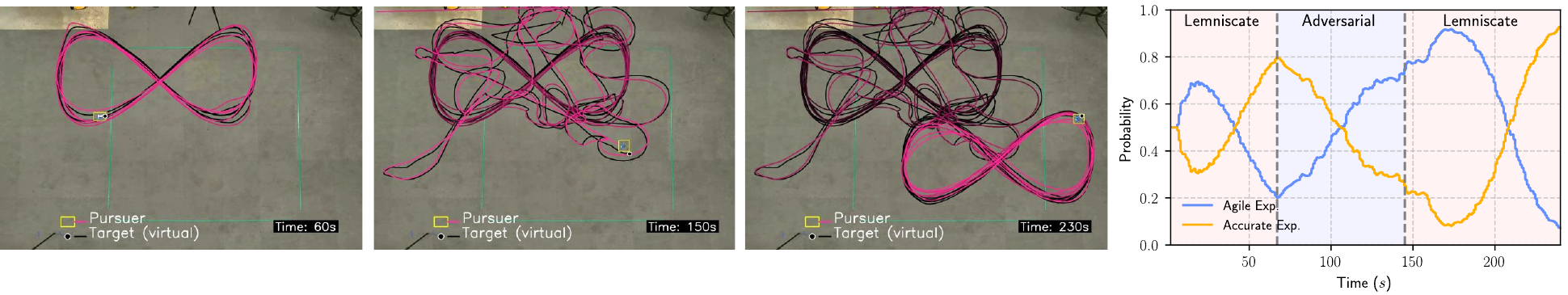}
    \text{\footnotesize \hspace{18pt}(a) Snapshot at $t=60s$ \hspace{33pt} (b) Snapshot at $t=150s$ \hspace{33pt} (c) Snapshot at $t=230s$  \hspace{20pt} (d) Online adaptation of selection probability. }

    \caption{{\textbf{Simplified Motivating Example of Self-Adaptive Learning and Model Predictive Control for Tracking Targets with Unknown Dynamics.} We investigate a self-adaptive learning and predictive control method for tracking unknown dynamics, where a robot is required to predict and track \textit{a priori} unknown and non-stationary trajectories. For example, in (a)-(c) a flying robot (Crazyflie) needs to track a mobile target that exhibits time-varying behavior, specifically, a mixture of structured and adversarial motion. We provide a self-supervised method that simultaneously learns multiple prediction models from scratch, without prior training, and adaptively selects the best one to match the observed target behavior.  In (d), our algorithm adapts the probabilistic policy among two employed predictors, where the target starts with a deterministic trajectory, transitioning later to a randomized adversarial behavior. Specifically, the \textit{agile} expert shows a fast learn-and-forget capability that helps in tracking oblivious or adversarial targets. On the contrary, the \textit{accurate} expert shows a slow but richer learning capability that helps in tracking targets executing a smooth trajectory. Our method allows for simultaneous learning and adaptation to the best available predictor. \textbf{Hardware experiments with two physical Crazyflies~(Fig.~\ref{fig:hw_snapshot}) and four predictors are presented in \Cref{sec:results}.}}}
    \label{fig:hardware_p1}
\end{figure*}

\section{Introduction}\label{sec:intro}
Mobile robots are increasingly used in dynamic environments to automate complex tasks such as target tracking~\citep{chen2016tracking}, navigation through traffic~\citep{salzmann2020trajectron}, and mapping of dynamic environments~\citep{atanasov2015decentralized}. 
All such tasks require accurate and efficient prediction and tracking of dynamic objects.
Achieving both, accuracy and efficiency is challenging since these tasks often require the robots to predict the motion of objects, humans, etc., whose dynamics are \textit{a priori} unknown and non-stationary, \ie may exhibit structured, random, switching, or adversarial behaviors, or even a mix of those.
For example, pursuit-evasion requires the robots to track evading targets of an unknown mixture of maneuvering capabilities;  collision-free navigation requires the robots to predict the motion of pedestrians, other vehicles, etc., each typically with different intentions (trajectories); and mapping of dynamic environments often requires tracking moving landmarks with various unknown heterogeneous dynamics.

Current approaches for prediction and tracking of unknown dynamics typically employ
(i) robust and adaptive control~\citep{lavretsky2012robust,zhou1998essentials,mayne2005robust,mayne2011tube,slotine1991applied,krstic1995nonlinear,ioannou1996robust},
(ii) offline learning~\citep{salzmann2020trajectron,chen2022scept,yuan2021agentformer,alahi2016social,gupta2018social,sampedro2018image,luo2019end,zhao2021hierarchical,dionigi2023exploring,dionigi2024d}, or
(iii) online learning~\citep{agarwal2019online,foster2020logarithmic,zhang2022adversarial,adib2023online,karapetyan2023online,abbasi2014tracking,zhao2022non,tsiamis2024predictive,zhou2025simultaneous}.
Robust control can be conservative since it accounts for the worst-case target behavior~\citep{zhou1998essentials,lavretsky2012robust}, instead of planning based on a model for the target dynamics. 
Offline learning can be expensive as it requires collection of data for training~\citep{pierson2017deep}. They often rely on large data-sets and robust training to generalize to unseen target dynamics, instead of learning on-the-fly based on real-time observations. For this reason, these methods can lack adaptability to out-of-distribution dynamics~\citep{mohri2018foundations,abu2012learning}.
Adaptive control and online learning control methods typically react to the observed motion or rewards, rather than anticipating it~\citep{slotine1991applied,krstic1995nonlinear,ioannou1996robust,agarwal2019online,foster2020logarithmic,zhang2022adversarial,abbasi2014tracking,zhao2022non,adib2023online,karapetyan2023online}.
This may lead to suboptimal performance in comparison to the prediction-based planning using a learned target.
State-of-the-art online learning for predictive control methods, such as~\cite{tsiamis2024predictive, zhou2025simultaneous}, require tuning tied to fixed target behavior, being unable to self-adapt to non-stationary target dynamics. {For example,~\cite{zhou2025simultaneous} introduces online least-squares via random Fourier features~(\RFF) for system identification. While this method is shown to rapidly learn fixed unknown dynamics on-the-fly, it lacks adaptability to switching unknown dynamics. Online deep learning methods~\citep{valkanas2025modl,sahoo2018online} have enabled neural networks to learn using sequential online data. While effective when large-scale datasets are available, these methods have limitations in real-time mobile robotics and lack guarantees.} 
We further discuss related works in \Cref{sec:lit_review}.

\begin{figure*}[t]
    \centering
    \includegraphics[width=\linewidth]{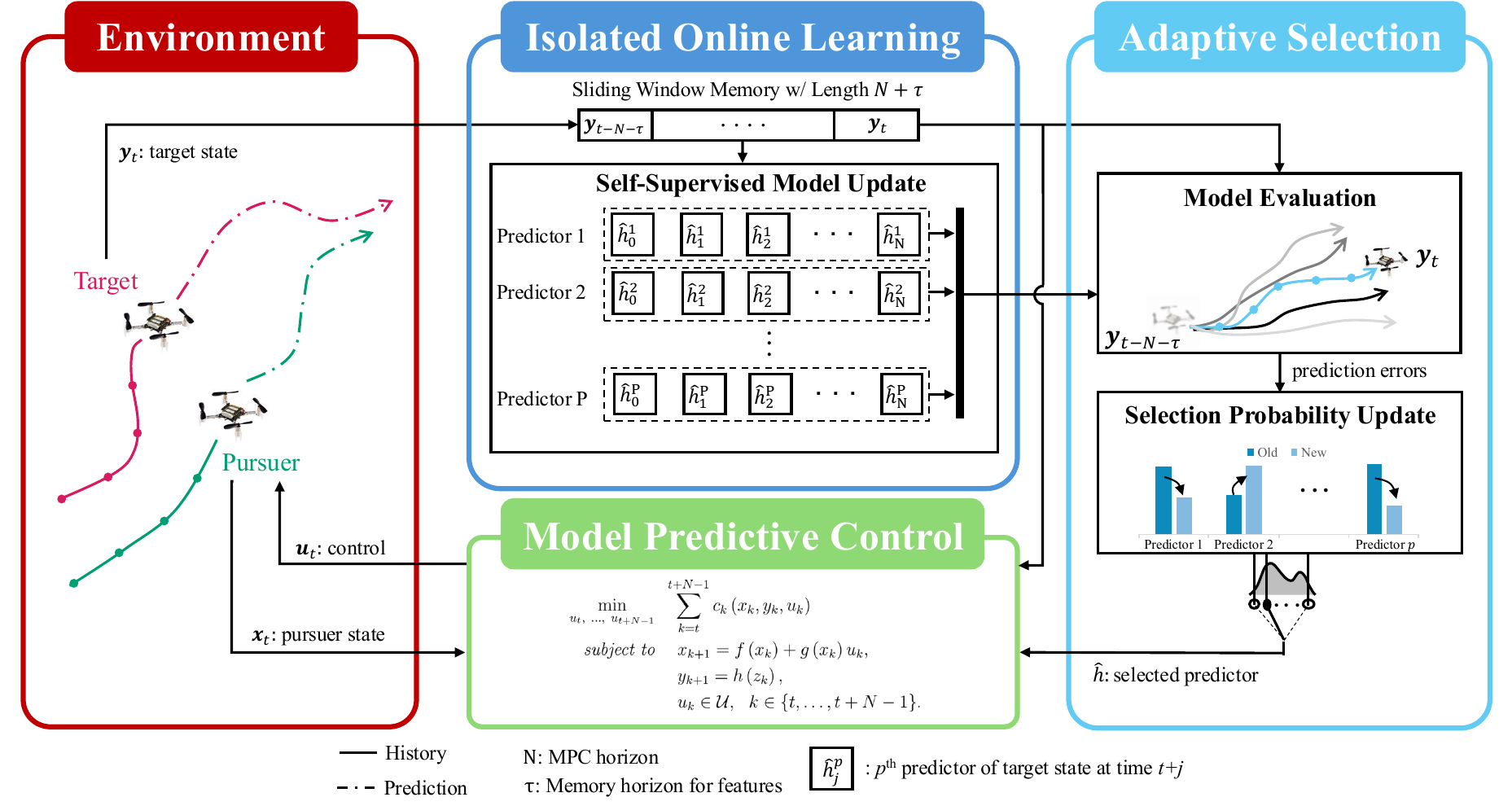}
    \caption{\textbf{Self-Adaptive Learning and \MPC for Tracking Unknown Dynamics.} The method composes three interacting modules: (i) isolated (self-supervised) online learning of target dynamics~(\ISO), (ii)  adaptive selection of predictors~(\AS), and (iii) model predictive control~(\MPC).  In summary, \ISO trains on-the-fly multiple predictors, \AS evaluates the current best predictor, and \MPC uses the selected best predictor by \AS to approximate the unknown dynamics and compute the control. }
    \label{fig_framework}
    \vspace{-3mm}
\end{figure*} 

\myParagraph{Contributions} 
In this paper, we provide a real-time, self-adaptive, and near-optimal online learning for control method for predicting and tracking unknown target dynamics, where the target dynamics can exhibit non-stationary behavior, 
\ie a mixture of structured and random motion~(Fig.~\ref{fig_framework}).
To this end, we unify (i) expert algorithms for self-adaptive learning~\citep{lattimore2020bandit,herbster1998tracking} and (ii) model predictive control for accurate control~\citep{rawlings2017model,borrelli2017predictive}.  
The method simultaneously learns multiple predictors from scratch, without prior training, and adaptively selects the best predictor to match the observed target behavior.
Therefore, the proposed method enables (i) one-shot online learning, (ii) control planned over a look-ahead horizon based on predictive models of the unknown dynamics, (iii) and online adaptation to switching unknown dynamics.
With this work, we do {not} profess that training-based offline learning is redundant; instead, we focus on the need for rapid online adaptation when robots are deployed in highly unknown and dynamic environments.
We provide sub-optimality guarantees in terms of regret, which captures the difference between the cumulative cost incurred by the proposed method and that of an optimal non-causal controller that knows the target dynamics \textit{a priori}.
We prove that the proposed method enjoys finite-time near-optimality guarantees in expectation, characterized as a function of the learning error of the target dynamics and the frequency that the target dynamics switch.
In the absence of both error and switching, our method asymptotically matches the optimal controller, achieving vanishing average regret over time. In the presence of only learning error, performance degrades smoothly with the magnitude of the prediction error, remaining close to optimal when errors are small. 
In the presence of both error and switching, the performance further degrades gracefully with the frequency of switching, with only sublinear dependence on the number of switches, demonstrating robustness to rapidly changing. Formal statements of these guarantees are provided in \Cref{sec:regret}.
All the regret guarantees hold true assuming boundedness of the robot and target states, Lipschitzness of the value function, and Lipschitzness of the control input~(Assumptions \ref{assump:bounded_state}-\ref{assump:perturbation}).
The analysis requires that the target dynamics do not depend on the robot. In the experiments, we also evaluate the method against adversarial targets that react to the robot~(\Cref{sec:results}).

The proposed method is composed of three interacting modules: (i) isolated (self-supervised) online learning of target dynamics~(\ISO), (ii)  adaptive selection of predictors~(\AS), and (iii) model predictive control~(\MPC).  At each time step, \ISO trains on-the-fly multiple predictors to approximate the target dynamics, \AS evaluates the performance of all predictors and selects the best predictor, and \MPC uses the selected best predictor by \AS and computes the current control input. 
In more detail, the \ISO module maintains multiple independent predictors of unknown target dynamics.
Each predictor parametrizes the unknown dynamics via linear combinations of finite numbers of \RFF~\citep{rahimi2007random,rahimi2008uniform,brault2016random,minh2016operator}. 
These parameters are rapidly updated online using an efficient online gradient descent subroutine that is based on~\cite{zhou2025simultaneous}. 
The \AS module leverages a bandit learning algorithm \Hedge~\citep{lattimore2020bandit} with Fixed-Share update~\citep{herbster1998tracking} to select a suitable predictor at each time step from a given set of multiple predictors trained by \ISO. 
Unlike standard \Hedge with Fixed-Share update, the \AS module has an information delay \citep{joulani2016delay,korotin2020adaptive} since the future target states are unknown, and during which, each predictor (or expert) evolves simultaneously resulting in the parameter drift. This requires additional theoretical analysis to prove the performance and regret guarantees, which we present as part of this work.

\myParagraph{Comparison with our prior work~\cite{zhou2025simultaneous}}
Our previous work~\citep{zhou2025simultaneous} proposes an online learning framework, based on \RFF representation~\citep{boffi2022nonparametric}, for online system identification for adaptive \MPC given a reference trajectory. Instead, herein we consider a different problem that requires a novel algorithmic design and theoretical analysis. 
On the theoretical side, we consider a target tracking setting with an unknown and non-stationary target reference trajectory, instead of a given reference trajectory as assumed in~\cite{zhou2025simultaneous}.  Thus, tackling the new problem herein requires a new set of assumptions and a novel proof strategy. In particular, our analysis is based on the Lipschitz continuity of the value function and the \MPC policy, whereas~\cite{zhou2025simultaneous} relies on exponential stability assumptions for any learned model parameters. As a result, the regret guarantees derived in this work cannot be obtained as a direct extension of~\cite{zhou2025simultaneous}.
On the algorithmic side, the proposed method uses the system identification method in~\cite{zhou2025simultaneous} as a subroutine, due to that method's practical runtime for on-the-fly system identification.  Specifically, herein, rather than learning a single model as in \cite{zhou2025simultaneous}, we design multiple target-state predictors under different learning configurations, such as learning rates, kernel parameters, and history lengths of target information. To enable practical runtime for on-the-fly learning of each such predictor, each predictor is built from multiple single-step learners,  thus enabling on-the-fly training with the system identification method in~\cite{zhou2025simultaneous}. 
In our evaluations across different target types, a well-tuned \SSI method~\citep{zhou2025simultaneous} either matches or has up to $30\%$ worse tracking error compared to our method.
The predictor design also results in more stable and less jitter in target trajectory prediction than~\cite{zhou2025simultaneous}.
The practical benefit of our approach is illustrated in simulated and hardware experiments against state-of-the-art methods, including optimization- and neural-network-based methods, as we describe next.  

\myParagraph{Experiments} 
We validate the algorithm in physics-based Gazebo simulations and Crazyflie hardware for various target tracking scenarios.
In \Cref{subsec:sim_results}, we consider non-switching tracking target trajectories, including lemniscate, Lissajous, sinusoidal, star, random, and adversarial and compare our algorithm with three diverse state-of-the-art online learning algorithms: \PLOT~\citep{tsiamis2024predictive}, an online non-stochastic learning algorithm with regret-based guarantees; \SSI, which uses an \RFF-based efficient online gradient descent method from~\cite{zhou2025simultaneous} to train a single-step predictor, used recursively to produce the needed multi-step predictions in \MPC; and \MODL~\citep{valkanas2025modl}, which uses multiple neural networks for online training.  
In \Cref{subsec:hardware}, we then deploy our algorithm on the hardware using two Crazyflie drones~(\Cref{fig:hw_snapshot}) and consider switching target tracking dynamics~(\Cref{fig:hardware}). We consider (i) a target that starts with a smooth and periodic lemniscate trajectory, then changes to a randomized adversarial trajectory, and again switches back to a lemniscate trajectory, (ii) a target that first follows a star-shaped trajectory with sharp corners and then switches to a smooth Lissajous curve trajectory, and (iii) a target following the shape of the alphabetical letters "UM". 
Our method achieves better tracking performance, being able to self-adapt to the different target behaviors without further tuning across the scenarios. 
Although baselines with a fixed parameter setting can match our performance on one trajectory type, they perform poorly on others.
Across the tested targets in simulation and hardware, the second-best method, which is different for different targets, always has about $3-35\%$ higher tracking error than our method.
Averaged across all the targets, any single baseline has at least $38\%$ higher average tracking error than our method in simulation and hardware.
In \Cref{subsec:indep_comp}, we analyze the individual components of the proposed method. Particularly, we isolate the individual modules, \ISO and \AS, to study the effects of various parameters and conditions that can vary for different target trajectory types.

\myParagraph{Structure of the Paper}
We discuss related work in \Cref{sec:lit_review}.
We define the problem of tracking unknown dynamics with \MPC in \Cref{sec:problem}. 
We elaborate on the general method composed of the \ISO and \AS modules in \Cref{sec:framework}. 
We describe our online optimization approach for the \ISO module in~\Cref{sec:online} and bandit learning for the \AS module in~\Cref{sec:selection}.  We present the theoretical performance guarantees in~\Cref{sec:regret}.
We provide simulation results, hardware experiment results, and ablation studies on \ISO and \AS modules in~\Cref{sec:results}. 
Summary and future directions are discussed in \Cref{sec:con}.
Proof, empirical regret results, and algorithm implementation details are given in the Appendix~(\Cref{sec:app}).

\section{Related Work}\label{sec:lit_review}

Methods for optimal decision making in hard-to-predict environments have been studied across control, operations research, and machine learning, including methods for offline learning, robust and adaptive control, online learning for control, as well as meta-learning and expert algorithms:

\paragraph{Offline learning for prediction and tracking.}
These works rely on neural networks that require collecting data to explicitly predict unknown trajectories into the
future~\citep{salzmann2020trajectron,chen2022scept,yuan2021agentformer,alahi2016social,gupta2018social} or to directly generate control input that enables tracking unknown trajectories~\citep{sampedro2018image,luo2019end,zhao2021hierarchical,dionigi2023exploring,dionigi2024d}.
The offline data collection and training process can be expensive and time-consuming, and the trained models may not generalize to unseen environments~\citep{mohri2018foundations,abu2012learning}.
In contrast, our method requires no offline learning, and employs one-shot, self-supervised learning using only online-collected data to predict the unknown trajectories.

\paragraph{Robust and adaptive control.}
The problem of tracking unknown trajectories can be reformulated as a control problem under unknown disturbances~\citep{boffi2022nonparametric,karapetyan2023online,zhang2022adversarial} 
and, thus, can be tackled by robust and adaptive control.
Robust control algorithms account for the worst-case realization of the disturbances based on the upper bound to their magnitude~\citep{lavretsky2012robust,zhou1998essentials,mayne2005robust,mayne2011tube}.
However, this can often be conservative, compromising control accuracy. 
Similarly, adaptive control compensates for the estimated disturbances based on retrospective data
~\citep{slotine1991applied,krstic1995nonlinear,ioannou1996robust}, instead of learning a predictive model of the disturbances/trajectories to enable receding horizon control.

\paragraph{Online learning for control.} 
The online learning for control algorithms~\citep{agarwal2019online,foster2020logarithmic,zhang2022adversarial,adib2023online,karapetyan2023online,abbasi2014tracking,zhao2022non,tsiamis2024predictive,zhou2025simultaneous} quantify the performance of the algorithms through \textit{regret}, \ie the suboptimality against an optimal clairvoyant controller that knows the unknown disturbances/trajectories. 
They provide bounded regret guarantees (i) upon employing the online convex optimization framework to update online the control input~\citep{agarwal2019online,foster2020logarithmic,abbasi2014tracking,zhang2022adversarial,adib2023online,karapetyan2023online,zhao2022non} or (ii) upon employing the recursive least squares to update the model of the unknown~\citep{tsiamis2024predictive,zhou2025simultaneous}.
However, \cite{agarwal2019online,foster2020logarithmic,abbasi2014tracking,zhang2022adversarial,adib2023online,karapetyan2023online,zhao2022non} update an auxiliary control input based on observed disturbances/trajectories instead of building a predictive model, thus becoming highly sensitive to tuning parameters and not able to handle fast-changing disturbances/trajectories.
Compared to \cite{zhou2025simultaneous}, we consider a target tracking setting with an unknown and non-stationary target reference trajectory, instead of a given reference trajectory as assumed in~\cite{zhou2025simultaneous}.
Thus, tackling the new problem herein requires a new set of assumptions and a novel proof strategy.
Particularly, our analysis is based on the Lipschitz continuity of the value function and the \MPC policy, whereas~\cite{zhou2025simultaneous} relies on exponential stability assumptions for any learned model parameters.
\cite{tsiamis2024predictive} adopts an \ISO-like \MPC framework for predicting target trajectories under linear robot and target dynamics, as opposed to the nonlinear dynamics and predictors we focused on herein. In addition, lacking an \AS module, \cite{tsiamis2024predictive} requires parameter tuning for targets with different levels of adversariality, as observed in \Cref{sec:results}. In contrast, we introduce a mixture-of-experts algorithm (the \AS module) to enable automatic selection of the best parameters on the fly.

\paragraph{Meta-learning and mixture of experts.}
Meta-learning and mixture of experts algorithms have drawn significant attention across operations research, machine learning, and control as effective strategies for adapting to nonstationary environments~\citep{lattimore2020bandit,masoudnia2014mixture,mu2025comprehensive}.
For example, \cite{chua2018deep,jesawada2025dr} use an ensemble of probabilistic dynamic models to provide uncertainty and data efficiency for model-based reinforcement learning. While it significantly reduces the number of trials in episodic learning, it is not directly suitable for scenarios where target policy may change online or the data is available sequentially, unlike batch training. 
Online deep learning methods such as~\cite{sahoo2018online,valkanas2025modl} use multiple neural networks and modular architecture for online backpropagation, but require large amounts of data to converge and lack adaptability for out-of-distribution samples.
Meta-learning methods build neural-network models that can rapidly adapt to new tasks or environments through online updates, enabling fast transfer across different conditions~\citep{finn2017model,nagabandilearning,richards2023control}.
Multi-task learning~\cite{caruana1997multitask,zhang2018overview,thung2018brief} is designed to learn diverse skills and enhance their ability to handle more complex tasks, such as complicated loco-manipulation~\citep{kuang2025skillblender,huang2025moe} and aggressive flight~\citep{xing2024multi}.
Both meta-learning and multi-task learning methods require offline training, thus exhibiting similar drawbacks to the offline learning methods.
In this paper, we exploit bandit algorithms~\citep{lattimore2020bandit} that, thus far, have been mainly employed in operations research to select the best-performing expert with guarantees in domains such as clinical trials~\citep{durand2018contextual}, recommendation systems~\citep{silva2022multi}, anomaly detection~\cite{ding2019interactive}, and wireless monitoring~\citep{le2014sequential}.
Applications of bandit algorithms in the field of control and robotics include trajectory forecasting~\citep{tong2025online}, multi-robot coordination~\citep{xu2023bandit},
system stabilization~\citep{li2023online}, disturbance rejection~\citep{zhao2022non}, power system control~\citep{feng2025online}, deformable object manipulation~\citep{mcconachie2020bandit}, and sampling-based motion planning~\citep{faroni2023motion}.

\section{Tracking Unknown Dynamics \\ With Model Predictive Control}\label{sec:problem}

In this section, we define the problem of {tracking unknown dynamics with model predictive control}. In particular, we consider our robot with control-affine dynamics:
\begin{equation}
	x_{t+1} = f\left(x_{t}\right) + g\left(x_{t}\right) u_{t}  , \quad t \geq 1, 
    \label{eq:affine_sys}
\end{equation}
where $x_t \in\mathbb{R}^{d_x}$ is the robot state, $u_t \in\mathbb{R}^{d_u}$ is the control input, $f: \mathbb{R}^{d_x} {\rightarrow} \mathbb{R}^{d_x}$ and $g: \mathbb{R}^{d_x} {\rightarrow} \mathbb{R}^{d_x} \times \mathbb{R}^{d_u}$ are known locally Lipschitz functions. Also, we consider \textit{target dynamics} unknown to the robot, to be tracked:
\begin{equation}
	y_{t+1} = h\left(z_{t}\right)  , \quad t \geq 1, 
    \label{eq:target}
\end{equation}
where $y_t \in\mathbb{R}^{d_y}$ is the target state, assumed bounded for all $t$, $h: \mathbb{R}^{d_z} \rightarrow \mathbb{R}^{d_y}$ is an unknown locally Lipschitz function, and $z_t \in\mathbb{R}^{d_z}$ is a vector of features chosen as a history of target state $y_{t-\tau:t}$. 
The history window $\tau$ is a hyperparameter chosen such that the past information is sufficient for predicting $y_{t+1}$.
{A high-level intuition and guidance for choosing $\tau$ is discussed in \Cref{app:impl_details}.}
$h$ implicitly depends on time $t$ in that the target dynamics may be switching behavior, \eg from deterministic to random, or a mixture of them.  
Notwithstanding, the theoretical analysis requires that the target dynamics do not depend on the robot. In the experiments, we also evaluate the method against adversarial targets that react to the robot~(\Cref{sec:results}).

Model predictive control (\MPC) selects a control input $u_t$ by simulating the robot dynamics and target motion over a look-ahead horizon $N$~\citep{rawlings2017model,borrelli2017predictive}: 
\begin{subequations}
    \label{eq:mpc_def}
    \begin{align}
        &\hspace{-4mm}\underset{{u}_{t}, \ \ldots, \ {u}_{t+N-1}}{\textit{min}} \hspace{4mm}\sum_{k=t}^{t+N-1} c_{k}\left(x_{k},y_{k},u_{k}\right) \label{eq:mpc_def_obj} \\
        & \ \ \operatorname{\textit{subject~to}} \;\quad x_{k+1} = f\left(x_{k}\right) + g\left(x_{k}\right) u_{k},\\
        & \qquad \qquad \qquad \; y_{k+1} = h\left(z_{k}\right), \\
        & \qquad \qquad \qquad \; u_{k}\in \calU, \ \ k\in\{t,\ldots, t+N-1\},
    \end{align}
\end{subequations}
where $c_{t}\left(\cdot,\cdot,\cdot\right): \mathbb{R}^{d_{n}} \times \mathbb{R}^{d_{n}} \times \mathbb{R}^{d_{u}} {\rightarrow} \mathbb{R}$ is the cost function of tracking performance and control effort, and $\calU$ is a compact set that represents constraints on the control input due to, \eg controller saturation.
We consider $c_{t}\left(x_t,y_t,u_t\right) \triangleq \|e_t\|^2_{Q}+\|u_t\|^2_{R}$, where $e_t \triangleq x_t - y_t$ is the tracking error, and $Q \in \mathbb{R}^{d_{n} \times{d_{n}}} \succeq 0$ and $R \in \mathbb{R}^{d_u \times d_{u}}\succ 0$ are weight matrices.

The optimization problem in \cref{eq:mpc_def} requires knowing $h\left(\cdot\right)$. In this paper, we propose a method to estimate $h\left(\cdot\right)$ online to handle the case where $h\left(\cdot\right)$ is unknown.  
Thus,  \cref{eq:mpc_def} is adapted to the optimization problem
where $c_{k}\left(x_{k},y_{k},u_{k}\right)$ and $y_{k+1} = h\left(z_{k}\right)$ are replaced by $c_{k}\left(x_{k},\hat{y}_{k},u_{k}\right)$ and $\hat{y}_{k+1} = \hat{h}\left(z_{k}\right)$.
We assess the control performance of our algorithm in comparison with a non-causal optimal controller that knows the actual target dynamics $h(\cdot)$ \textit{a priori}.

\begin{definition}[Dynamic Regret for Tracking]\label{def:DyReg_control}
Given a total time horizon of length $T$, and corresponding loss functions $c_t$, $t=1,\ldots, T$, the \emph{dynamic regret} is defined as
\begin{equation}
	\DReg \triangleq J_{T}\left(\pi; y_{1:T}\right)-J_{T}\left(\pi^\star; y_{1:T}\right),
	\label{eq:DyReg_control}
\end{equation}
where $J_{T}\left(\pi; y_{1:T}\right) \triangleq \sum_{t=1}^{T} c_{t}\left(x_{t}, y_{t}, u_{t}\right)$, $J_{T}\left(\pi^\star; y_{1:T}\right) \triangleq \sum_{t=1}^{T} c_{t}\left(x_{t}^\star, y_{t}, u_{t}^\star\right)$, $\pi$ is the \MPC policy in \cref{eq:mpc_def} that calculates $u_t$ based on $\hat{h}\left(\cdot\right)$,  
$\pi^\star$ is the \MPC policy in \cref{eq:mpc_def} with known target dynamics ${h}\left(\cdot\right)$,
$x_{t}^{\star}$ is the optimal robot state reached by $u_{t}^{\star}$.
\end{definition}

\begin{problem}[Tracking Unknown Dynamics With Model Predictive Control]\label{prob:tracking}
Suppose $x_1$ and $y_1$ are known, and $x_t$ and $y_t$ can be measured.
At each time-step $t$, update $\hat{h}\left(\cdot\right)$, and identify a control input such that the regret $\DReg$ is bounded. 
\end{problem}

\section{Overview of Method for Self-Adaptive Online Learning for Control}\label{sec:framework}

We overview the proposed method for self-adaptive learning and \MPC for tracking unknown dynamics.  The method is depicted in \Cref{fig_framework}.
Particularly, the method is composed of three interacting modules: (i) isolated (self-supervised) online learning of target dynamics~(\ISO), (ii)  adaptive selection of predictors~(\AS), and (iii) model predictive control~(\MPC).  
The role of each module is as follows: 
\begin{itemize}
    \item  \ISO learns on-the-fly multiple target predictors, in a self-supervised manner, employing different types of motion models and learning parameters. To this end, the predictors are trained based on the sliding-window history of observed target states $y_{t-N-\tau:t}$.  
    \item \AS adaptively selects the current best predictor $\hat{h}$ for the \MPC module. To this end, \AS evaluates each predictor's performance based on the history $y_{t-N-\tau:t}$ and updates a probabilistic selection policy over the predictors.
    \item \MPC employs the selected predictor by \AS to predict the unknown target dynamics and, thus, calculate the next control input for the robot.
\end{itemize}

\section{Isolated Online Learning}\label{sec:online}
We present the \ISO algorithm (\Cref{alg:expert}),
with Fig.~\ref{fig_framework} presenting an overview of the internal structure of \ISO. The module maintains $P$ independent trajectory predictors for the unknown target dynamics. Each predictor is characterized by a set of hyperparameters for learning.  Each set of hyperparameters is chosen to enable learning of a particular anticipated target type. Particularly, each predictor is composed of $N$ learners $\hat{h}_{j}^{p}$ that learn the target dynamics for each step in the \MPC horizon~(\Cref{subsec:multiple}). 
\textit{For a specific $\hat{h}_{j}^{p}$, the learning update subroutine is based on random approximation of motion models using \RFF~\citep{boffi2022nonparametric}~(\Cref{subsec:RFF}). The model parameters are rapidly updated online via online least-squares (on the past target state data as features) per the efficient online gradient descent approach in~\cite{zhou2025simultaneous} (\Cref{subsec:OLS}).}
The tuning of hyperparameters is discussed in \Cref{app:impl_details}. 

\subsection{Predicting with Multiple Motion Models}\label{subsec:multiple}
The unknown target motion is represented by a single motion model $\hat{h}\left(\cdot\right)$ 
in \Cref{sec:problem}.
However, predicting future target states recursively through a single motion model can lead
to instability~\citep{tsiamis2024predictive}, \eg due to the accumulation of the estimation error. 
Instead, we herein utilize multiple target motion models to predict future target states.
We show that a single motion model can be replaced by multiple target motion models for multi-step prediction, that is, for predicting the future target states in \MPC. To this end, let $\hat{y}$ be the simulated target state in \MPC. By the definition of $z_t$, 
\begin{equation}
    \begin{aligned}
        \hat{y}_{t+1} &= \hat{h}\left(z_{t}\right) \triangleq \hat{h}_{0}\left(y_{t-\tau}, \; \dots, \; y_{t}\right),  \\
        \hat{y}_{t+2} &= \hat{h}\left(z_{t+1}\right) = \hat{h}_{0}\left(y_{t-\tau+1}, \; \dots, \; \hat{y}_{t+1}\right)  \\
        &\qquad \qquad \; \; \; \triangleq \hat{h}_{1}\left(y_{t-\tau}, \; \dots, \; y_{t}\right), \\
        & \; \; \vdots  \\
        \hat{y}_{t+N} &= \hat{h}\left(z_{t+N-1}\right) = \hat{h}_{0}\left(y_{t-\tau+N-1}, \; \dots, \; \hat{y}_{t+N+1}\right)  \\
        &\qquad \qquad \qquad  \triangleq \hat{h}_{N-1}\left(y_{t-\tau}, \; \dots, \;y_{t}\right),  
    \end{aligned}
    \label{eq:multi_target_model}
\end{equation}
where we make explicit the dependence of $\hat{h}_{j}\left(\cdot\right)$ on feature vector $\tilde{z}^{}\triangleq \left[y_{t-\tau}^\top, \; \dots, \;  {y}_{t}^\top\right]^\top$, and $j \in \{0, \dots, N-1\}$.
Therefore, we can adopt \MPC in \cref{eq:mpc_def} to the following with multiple target motion models.

\begin{subequations}
    \label{eq:mpc_ada_multi_def}
    \begin{align}
       &\hspace{-4mm}\underset{{u}_{t}, \ \ldots, \ {u}_{t+N-1}}{\textit{min}} \hspace{4mm} \sum_{k=t}^{t+N-1} c_{k}\left(x_{k},y_{k},u_{k}\right) \label{eq:mpc_ada_obj} \\
        & \ \ \operatorname{\textit{subject~to}} \;\quad x_{k+1} = f\left(x_{k}\right) + g\left(x_{k}\right) u_{k}, \label{eq:mpc_ada_dyn}\\
        & \qquad \qquad \qquad \; y_{k+1} = \hat{h}_{j}\left(\tilde{z}\right), \label{eq:mpc_ada_tar} \\
        & \qquad \qquad \qquad \ u_{k}\in \calU, \ \ j = k-t, \\
        & \qquad \qquad \qquad \ \tilde{z}^{}\triangleq \left[y_{t-\tau}^\top, \; \dots, \;  {y}_{t}^\top\right]^\top, \\
        & \qquad \qquad \qquad  \ k\in\{t,\ldots, t+N-1\}, 
    \end{align}
\end{subequations}
where $\hat{h}_{j}\left(\cdot\right)  \triangleq \hat{h}_{j}\left(\cdot~; \hat{\alpha}^{j}\right)$ is parameterized by $\hat{\alpha}^{j}$ that is updated online.

\subsection{RFF Parametrization of Motion Models}\label{subsec:RFF}

We describe the approach for learning $h_{j}\left(\cdot\right)$ via finite numbers of randomly sampled Fourier features (\RFF)~\citep{rahimi2007random,rahimi2008uniform,brault2016random,minh2016operator}.
We use \RFF since they are able to approximate functions in Reproducing Kernel Hilbert Space~(\RKHS) with a provable uniform approximation error~(\Cref{prop:approx_error}). 

For simplicity, we omit the subscript $j$, and consider $h: \mathbb{R}^{d_z} \rightarrow \mathbb{R}^{d_y}$ to lie in a subspace of a \emph{\RKHS} $\calH$, where the kernel $K$ is considered known~\citep{carmeli2010vector} and is written via a feature map $\Phi: \mathbb{R}^{d_z} \times \Theta \rightarrow \mathbb{R}^{d_y \times d_1}$ as $K\left(z_1, z_2\right) = \int_\Theta \Phi\left( z_1, \theta \right) \Phi\left( z_2, \theta \right)^\top \mathrm{d} \nu(\theta),$
where $d_1 \leq d_y$, $\nu$ is a known probability measure on a measurable space $\Theta$~\citep{bach2017breaking}. 
Per~\cite{brault2016random},
the measurable space $\Theta\subset\mathbb{R}^{d_z+1}$, and the feature map of any translation-invariant kernel can be written as $\Phi \left(z, \theta\right)= B(w) \phi\left(w^{\top} z+b\right)$, where $\theta=\left(w, b\right)$, $\theta \in \Theta$, $w \in \mathbb{R}^{d_z}$ and $b \in \mathbb{R}$. The function $B: \mathbb{R}^{d_z} \rightarrow \mathbb{R}^{d_y \times d_{1}}$ and $1$-Lipschitz function $\phi: \mathbb{R} \rightarrow[-1,1]$.
Such kernels include the Gaussian and Laplace kernels.
Then, function $h$ can be written as $h\left(\cdot\right) = \int_\Theta \Phi\left(\cdot, \theta\right) \alpha(\theta) \mathrm{d}\nu (\theta)$,
with $\left\| h \right\|_\calH^2 = \left\| \alpha \right\|_{L_2\left(\Theta, \nu\right)}^2 \triangleq \int_\Theta \left\| \alpha(\theta) \right\|^2 \mathrm{d} \nu(\theta)$, where $\alpha: \Theta \rightarrow \mathbb{R}^{d_1}$ is a square-integrable signed density~\citep{bach2017breaking}. The corresponding Hilbert space is referred to as $\calF_2$~\citep{bach2017breaking,bengio2005convex}.

A finite-dimensional approximation of $h\left(\cdot\right)$ is obtained by
\begin{equation}
    \hat{h}(\cdot;\alpha)\triangleq\frac{1}{M} \sum_{i=1}^{M} \Phi\left(\cdot, \theta_i \right) \alpha_i,
    \label{eq:kernal_approx}
\end{equation}
where $\theta_i \sim \nu$ are drawn i.i.d. from the base measure $\nu$, $\alpha_i \triangleq \alpha\left(\theta_i\right)$ are {parameters that need to be learned, and $M$ is the number of sampling points, which determines the accuracy of the approximation, per the following proposition.}

\begin{proposition}[Uniformly Approximation Error
~\citep{boffi2022nonparametric} (simplified version)]\label{prop:approx_error}
     Assume $h \in$ $\mathcal{F}_{2}\left(B_{h}\right)$, where $\calF_2 \left(B_h\right) \triangleq  \Bigg\{ {h}\left(\cdot\right) = \left. \int_\Theta {\Phi}\left(\cdot, {\theta}\right) {\alpha}({\theta}) \mathrm{d}\nu ({\theta})  \right\vert {\alpha} \in \calD \Bigg\}$
     and $\calD\triangleq \{ {\alpha} \mid \|{\alpha}\| \leq B_h\}$.
     Then, with high probability, there exist $\left\{\alpha_{i}\right\}_{i=1}^{M} \in \calD$ such that
\begin{equation}
        \left\| h\left(\cdot\right) - \frac{1}{M} \sum_{i=1}^{M} \Phi\left(\cdot, \theta_{i}\right) \alpha_{i}\right\|_{\infty} 
        = 
        \calO\left(\frac{1}{\sqrt{M}}\right).
    \end{equation}
\end{proposition}

\subsection{Online Learning of RFF Parametrizations}\label{subsec:OLS}
We utilize online least-squares estimation to update the parameter $\alpha^{j}$ of each $\hat{h}_{j}$. 
Specifically, upon observing $y_{t+1}$, we leverage the history information 
$\tilde{z}_{j}\triangleq y_{t-1-\tau-j:t-1-j}$
and prediction $\hat{h}_{j}\left(\tilde{z}_{j}\right)$ to update the parameters $\hat{\alpha}_{t}^{j} \triangleq \left[ \alpha_{i,t}^{j\top}, \; \dots, \; \alpha_{M,t}^{j\top}\right]^\top$ and to minimize the approximation error $l_t^j = \| y_{t} - \hat{h}_{j}\left(\tilde{z}_{j}\right) \|^2$, where $ \hat{h}_{j}(\cdot) \triangleq  \frac{1}{M} \sum_{i=1}^{M} \Phi\left(\cdot, \theta_i^j \right) \hat{\alpha}_{i,t}^j $ and $\Phi\left(\cdot,\theta_i^j\right) $ is the random Fourier feature as in \Cref{subsec:RFF}. 
The algorithm uses the online gradient descent algorithm~(\OGD)~\citep{hazan2016introduction}. At each $t = 1, \dots, T$ for each $j$,
we first formulate the estimation loss function (approximation error) given $\left( \tilde{z}, \; y_{t+1} \right)$ as $l_{t}^{j}\left(\hat{\alpha}_t\right) \triangleq \left\| y_{t}-  \frac{1}{M} \sum_{i=1}^{M} \Phi\left(\tilde{z}_{j}, \theta_i^{j} \right) \hat{\alpha}_{i,t}^{j} \right\|^2$.
Then we calculate the gradient of $l_{t}^{j}\left(\hat{\alpha}_t\right)$ with respect to $\hat{\alpha}_{t}^{j}$ by $\nabla_{t}^{j} \triangleq \nabla l_{t}^{j}\left(\hat{\alpha}_{t}^{j}\right)$.
The parameters are updated by $\hat{\alpha}_{t+1}^{j \prime}= \hat{\alpha}_{t}^{j}- \eta \nabla_{t}^{j}$ with learning rate $\eta$.
Finally, project each $\hat{\alpha}_{i,t+1}^{j\prime}$ onto $\calD_{j}$ using $\hat{\alpha}_{i,t+1}^{j} = \Pi_{\calD_{j}}(\hat{\alpha}_{i,t+1}^{j\prime}) \triangleq \underset{\alpha \in \calD_{j}}{\operatorname{\textit{argmin}}}\; \| \alpha - \hat{\alpha}_{i,t+1}^{j\prime} \|^2$.

\begin{algorithm}[t]
\caption{{Isolated Online Learning for Predicting Unknown Trajectory (\ISO).}}
	\begin{algorithmic}[1]
		\REQUIRE Sliding window length $\tau$; number of random Fourier features $M$; base measure $\nu_{j}$; domain set $\calD_{j}$;  gradient descent learning rate $\eta$; history information window $\tau$; \MPC horizon $N$; $j \in \{0, \dots, N-1\}$.
		\ENSURE At each time step $t=1,\ldots,T$, learned target motion model $\hat{h}_{j}\left(\cdot\right)$.
		\medskip
            \STATE Initialize $x_1$, $y_1$, $\hat{\alpha}_{i,0}^{j} \in \calD_{j}$; 
            \STATE Randomly sample $\theta_i^{j} \sim \nu_{j}$ and formulate $\Phi\left(\cdot, \theta_i^{j}\right)$, where $i \in \{1, \dots, M\}$;
		\FOR {each time step $t = 1, \dots, T$}
                \STATE Observe target state $y_{t}$;
    		\FOR {each $j = 0, \dots, N-1$}
                \STATE Formulate feature vector $\tilde{z}_{j}\triangleq y_{t-1-\tau-j:t-1-j}$;
                \STATE Formulate estimation loss $l_{t}^{j}\left(\hat{\alpha}_t^j\right) \triangleq \left\| y_{t}-  \frac{1}{M} \sum_{i=1}^{M} \Phi\left(\tilde{z}_{j}, \theta_i^{j} \right) \hat{\alpha}_{i,t}^{j} \right\|^2$;
                \STATE Calculate gradient $\nabla_{t}^{j} \triangleq \nabla l_{t}^{j}\left(\hat{\alpha}_{t}^{j}\right)$;
                \STATE Update $\hat{\alpha}_{t+1}^{j \prime}= \hat{\alpha}_{t}^{j}- \eta \nabla_{t}^{j}$;
                \STATE Project $\hat{\alpha}_{i,t+1}^{j\prime}$ onto $\calD_{j}$, \ie $\hat{\alpha}_{i,t+1}^{j} = \Pi_{\calD_{j}}(\hat{\alpha}_{i,t+1}^{j\prime})$, for $i \in \{1, \; \dots, \; M\}$;
                \ENDFOR
            \ENDFOR
	\end{algorithmic}\label{alg:expert}
\end{algorithm}

\section{Adaptive Selection of Predictors}\label{sec:selection}
We present the \AS algorithm (\Cref{alg:meta}).
The \AS module leverages the expert learning algorithm \Hedge~\cite[Chapter~11]{lattimore2020bandit} and its Fixed-Share variant~\citep{herbster1998tracking} to select a suitable predictor from a given ensemble of $P$ predictors (trained by \ISO) at each time step. To this end, \AS uses a performance metric to evaluate each predictor with $N$-step information delay~(\Cref{subsec:eva}) and maintains a probabilistic selection policy over the $P$ predictors~(\Cref{subsec:prob}).
{The tuning of hyperparameters is discussed in \Cref{app:impl_details}. }

\subsection{Predictor Performance Evaluation}\label{subsec:eva}

At each time $t$, consider predictor $p$ with $\hat{h}^p_j\left(\cdot~; \hat{\alpha}^{j,p}_{t}\right)$ from an ensemble of $P$ independent predictors, where $j\in\{0,N-1\}$, $p\in\{1, \dots, P\}$. Since we deal with a single predictor, we omit the superscript $p$ for simplicity. 

We consider a sliding window of memory of target states $y_{t-N-\tau:t}$. Using $\hat{h}^p_j\left(\cdot~; \hat{\alpha}^{j,p}_{t}\right)$, we can compute the target state predictions from time steps $t-N+1$ to $t$ as follows 
\begin{equation}
    \begin{aligned}
        \hat{y}_{t-N+1} &= \hat{h}_{0}\left(y_{t-N-\tau:t-N}\right),  \\
        \hat{y}_{t-N+2} &= \hat{h}_{1}\left(y_{t-N-\tau:t-N}\right),  \\
        & \;\; \vdots\\
        \hat{y}_{t} &= \hat{h}_{N-1}\left(y_{t-N-\tau:t-N}\right),  \\
    \end{aligned}
    \label{eq:predcitions}
\end{equation}

Next, we compare the predicted values to the ground-truth to assess the predictor's performance. 
The most natural metric is through the root mean square error (RMSE) as follows:
\begin{equation}
    r_t = \bar{r} - \sqrt{\sum_{j=0}^{N-1} \|y_{t-N+1+j} -\hat{y}_{t-N+1+j} \|^2},
    \label{eq:rmse_theory}
\end{equation}
where $\bar{r}$ is the upper bound of prediction errors.

{We can remove dependence on $\bar{r}$ by using  $r_t =  \sqrt{\sum_{j=0}^{N-1} \|y_{t-N+1+j} -\hat{y}_{t-N+1+j} \|^2}$ as loss (instead of reward), and replacing $\exp(\lambda r_{t}^{p})$ with $\exp(-\lambda r_{t}^{p})$ in \Cref{alg:meta}}.

\begin{algorithm}[t]
\caption{\mbox{Adaptive Selection of Predictors (\AS).}}
    \begin{algorithmic}[1]
        \REQUIRE Number of predictors $P$; learning rate $\lambda$; share rate $\beta\in[0,1)$; \MPC horizon $N$.
        \ENSURE At each time step $t=1,\ldots,T$, chosen predictor $\tilde{p}$.
        \medskip
        \STATE Initialize $\Delta_{0}^{p}=1/P$, for all $p \in \{1, \dots, P\}$;
        \FOR{each time step $t=1,\dots,T$}
            \FOR{each predictor $p=1,\dots,P$}
                \STATE Retrieve a sliding window of memory of target states $y_{t-N-\tau:t}$;
                \STATE Obtain predictions $\{\hat{y}_k\}_{k=t-N+1}^{t}$ using $\hat{h}^p_j\left(\cdot~; \hat{\alpha}^{j,p}_{t}\right)$;
                \STATE Calculate prediction accuracy $r_t^p$;
                \STATE Update $\bar{\Delta}_{t}^{p} = \frac{\Delta_{t-1}^{p}\exp(\lambda r_{t}^{p})}{\sum_{q} \Delta_{t-1}^{q}\exp(\lambda r_{t}^{q})}$;
            \ENDFOR
            \STATE Update selection policy by
                $\Delta_{t}^{p} = (1-\beta)\,\bar{\Delta}_{t}^{p} + {\beta}/{P}$, for all $p \in \{1, \dots, P\}$;
            \STATE Sample predictor $\tilde{p} \sim \Delta_t$;
        \ENDFOR
    \end{algorithmic}\label{alg:meta}
\end{algorithm}

\subsection{Adaptive Selection Policy}\label{subsec:prob}
Based on the RMSE of each predictor, we utilized the \Hedge algorithm~\cite[Chapter~11]{lattimore2020bandit} and the Fixed-Share update~\citep{herbster1998tracking} with $N$-step information delay to update the policy for adaptive selection of $P$ predictors.
The selection policy is a probability distribution $\Delta_t$ over the predictors, initialized as the uniform distribution $\Delta_0^p=1/P$. Upon calculating $r_t^p$ for all predictor $p \in \{1, \dots, P\}$, we first perform the exponential-weights update
\begin{equation}
    \bar{\Delta}_{t}^{p} = \frac{\Delta_{t-1}^{p}\exp(\lambda r_{t}^{p})}{\sum_{q} \Delta_{t-1}^{q}\exp(\lambda r_{t}^{q})},
    \label{eq:hedge_update}
\end{equation}
where $\lambda > 0$ is the learning rate. For $\beta=0$, unrolling \cref{eq:hedge_update} gives $\Delta_{t}^{p} = \frac{\exp(\lambda S_{t}^{p})}{\sum_{q} \exp(\lambda S_{t}^{q})}$, where $S_{t}^{p} = S_{t-1}^{p} + r_{t}^{p}$ is the cumulative prediction accuracy with $S_{0}^{p} = 0$. 
Then, we mix the resulting distribution with the uniform distribution,
\begin{equation}
    \Delta_{t}^{p} = (1-\beta)\,\bar{\Delta}_{t}^{p} + \frac{\beta}{P},
    \label{eq:fixed_share}
\end{equation}
where $\beta\in[0,1)$ is the share rate. \Cref{eq:fixed_share} guarantees that each predictor keeps a selection probability of at least $\beta/P$. Hence, when the best predictor switches, the new best predictor can regain a high selection probability within a number of steps that does not depend on how long the previous predictor has been the best. 
For standard \Hedge and Fixed-Share algorithm, one should use $y_{t-\tau:t}$ to predict $y_{t+1:t+N}$. However, $y_{t+1:t+N}$ is unknown at time $t$. Therefore, \Cref{alg:meta} utilizes $y_{t-N-\tau:t-N}$ to predict $y_{t-N+1:t}$, causing a $N$-step information delay.
Finally, a predictor $\tilde{p} \sim \Delta_t$ is chosen for the next control cycle by sampling from the distribution $\Delta_t$, where $\tilde{p}\in\{1,\dots,P\}$. This way, the \Hedge algorithm maintains a probabilistic adaptive policy over multiple predictors.

\section{Performance Guarantees}\label{sec:regret}
We first prove that the proposed method achieves no dynamic regret under assumptions of (i) boundedness of robot and target states~(\Cref{assump:bounded_state}), (ii) Lipschitzness of value function~(\Cref{assump:lipschitz}), (iii) Lipschitzness of \MPC Control Input~(\Cref{assump:perturbation}), and (iv) negligible approximation error of the target dynamics while the target dynamics are also assumed \underline{non}-switching~(\Cref{assump:small_approx_error}). 
Then, we show that the regret gracefully degrades in the presence of approximation error in the target dynamics~(\Cref{cor:regret_with_error}) and switching target dynamics~(\Cref{cor:switching}).
\textit{From a technical contribution point of view, although the proposed approach leverages \RFF-based online learning and \MPC~\citep{zhou2025simultaneous}, a novel proof, with a different set of assumptions, is needed for \Cref{theorem:regret}.
}

\begin{assumption}[Boundedness of Robot  and Target State]\label{assump:bounded_state}
    The \MPC policy in \cref{eq:mpc_def} renders the system $e_{t}^\star = x_{t}^\star - y_{t}$ uniformly bounded, and the \MPC policy in \cref{eq:mpc_ada_multi_def} renders the system $e_{t} = x_{t} - y_{t}$ uniformly bounded.
\end{assumption} 

Since $\hat{y}_{t}$ and ${y}_{t}$ are uniformly bounded, we expect that $\pi$ and $\pi^\star$ are well-tuned and able to achieve uniformly bounded tracking errors with respect to the target. 
Empirical evidence of bounded tracking error of nonlinear \MPC can be found in~\cite{sun2022comparative}.
\Cref{assump:bounded_state} assumes that the tracking errors are bounded, whereas \cite[Assumption~1]{zhou2025simultaneous} assumes global asymptotic stability, implying that the tracking errors go to zero for any $y_t$.

\begin{definition}[Value Function, $Q$-Function, and Advantage Function]
    Let $\pi_1$ and $\pi_2$ be two control policies, the value function $\boldsymbol{V}_t^{\pi_1}(e ; y_{1:T})$ is the cost-to-go by applying policy $\pi_1$, and the $Q$-function $\boldsymbol{Q}_t^{\pi_1}(x, u^{\pi_2} ; y_{1:T})$ is the cost-to-go by applying policy $\pi_2$ at time $t$ and policy $\pi_1$ afterwards. Specifically, they are defined as
    \begin{equation}
        \begin{aligned}
            \boldsymbol{Q}_t^{\pi_1}(x, u^{\pi_2} ; y_{1:T}) &= \|x - y_{t}\|_{Q}^2+\|u^{\pi_2}\|_{R}^2 + \\
            & \qquad \boldsymbol{V}_{t+1}^{\pi_1}\left(f\left(x\right) + g\left(x\right) u^{\pi_2} - y_{t+1}; y_{1:T}\right), \\
            \boldsymbol{V}_t^{\pi_1}(e ; y_{1:T})  &= \boldsymbol{Q}_t^{\pi_1}\left(x, u^{\pi_1} ; y_{1:T}\right) .
        \end{aligned}
    \end{equation}

    The advantage function for the $\pi_1$ over $\pi_2$ is 
    \begin{equation}
        \begin{aligned}
            \boldsymbol{A}_t^{\pi_1}(u^{\pi_2} ; x, y_{1:T}) &\triangleq \boldsymbol{Q}_t^{\pi_1}(x, u^{\pi_2} ; y_{1:T}) - \boldsymbol{Q}_t^{\pi_1}(x, u^{\pi_1} ; y_{1:T}) \\ 
            &= \boldsymbol{Q}_t^{\pi_1}(x, u^{\pi_2} ; y_{1:T}) - \boldsymbol{V}_t^{\pi_1}(e ; y_{1:T}),
        \end{aligned}
    \end{equation}
    representing the total excess cost incurred by selecting a control $u^{\pi_2}$ at state $x$ and time $t$, and following $\pi^{\pi_1}$ for the remaining time steps.
\end{definition}

We next assume the following Lipschitzness property of $\boldsymbol{V}_t^{\star}(e)$ with respect to $e$.  To this end, for notational simplicity, we set $\boldsymbol{Q}_t^{\star}(x, u) \triangleq \boldsymbol{Q}_t^{\pi^\star}(x, u^{\pi} ; y_{1:T})$, $\boldsymbol{V}_t^{\star}(e) \triangleq \boldsymbol{V}_t^{\pi^\star}(e ; y_{1:T})$ and $\boldsymbol{A}_t^{\star}(u ; x) \triangleq\boldsymbol{A}_t^{\pi^\star}(u^{\pi} ; x, y_{1:T})$. 

\begin{assumption}[Lipschitzness of Value Function]\label{assump:lipschitz}
    There exists a constant $L_V$, independent of $T$, such that $| \boldsymbol{V}_t^{\star}(e_1)- \boldsymbol{V}_t^{\star}(e_2)|\, \leq L_V \|e_1 - e_2 \|$, for all $e_1, e_2$ in a compact set and $t \in \{1, \dots,T\}$.
\end{assumption} 

\Cref{assump:lipschitz} is satisfied for, \eg the value function obtained by linear \MPC, \ie~\MPC with linear systems of the form $e_{t+1} = A e_{t} + B u_{t} + w_t$ and uniformly bounded $w_t$, assuming that sufficient control authority is available. In this case, we obtain $u_{t} = -K e_t - \sum_{k=t+1}^{t+N}K_{k-t} w_{k}$, where $K$ is Hurwitz, and $K_{k-t}$, $ w_{k}$ uniformly bounded~\citep{goel2022power,foster2020logarithmic}, and Lipschitzness of the value function can be analytically verified. Similarly, for nonlinear \MPC, \Cref{assump:lipschitz} is satisfied if \MPC is exponentially stable to zero reference (\ie setting $y$ as zero), which can be designed by \eg using sufficiently long \MPC horizon~\citep{grune2016nonlinear}, and $u_{t}$ is Lipschitz in the initial conditions $x_t$ and $y_{t+1}, \dots, y_{t+N}$, which can be verified by \eg checking strong second order sufficient conditions~\citep{dontchev2013euler,robinson1980strongly,dontchev2019lipschitz}.

We next assume the Lipschitz continuity of the \MPC input in the predicted target states, to quantify the difference between the control inputs given by \cref{eq:mpc_def} and \cref{eq:mpc_ada_multi_def} for different predictions of target states. 

\begin{assumption}[Lipschitzness of Control Input]\label{assump:perturbation}
    There exists a constant $L_\pi > 0$, independent of $t$ and $T$, such that, given the initial condition $x_t$ and $y_t$, the \MPC policies in \cref{eq:mpc_def} with ${y}_{t+1:t+N}$ and in \cref{eq:mpc_ada_multi_def} with $\hat{y}_{t+1:t+N}$ satisfy
    \begin{equation}
        \| u_{t} - u_{t}^{\star} \| \leq L_\pi \sum_{k=t+1}^{t+N} \|y_k -\hat{y}_k \|, \qquad t \in \{1,\dots,T\}.
        \label{eq:perturbation}
    \end{equation}
\end{assumption}

Lipschitz continuity of the control policy is a standard assumption in the regret analysis of predictive and online control, \eg in~\cite{lin2022bounded,karapetyan2025closed}, and it is milder than the exponentially decaying perturbation bounds established in~\cite{lin2021perturbation,xu2019exponentially,na2022superconvergence,shin2021controllability,shin2022exponential}. For the \MPC in \cref{eq:mpc_def}, \Cref{assump:perturbation} holds (i) globally for linear robot dynamics, \ie $f(x)=Ax$ and $g(x)\equiv B$, with a convex polyhedral $\calU$ (\eg a box), for which the \MPC problem is a strongly convex quadratic program since $R \succ 0$, and (ii) for nonlinear robot dynamics under strong second-order sufficient condition and linear independence constraint qualification for sufficiently small prediction errors~\citep{robinson1980strongly,dontchev2019lipschitz}, the latter of which holds in our experiments.

Finally, we assume there exists a predictor such that the uniform approximation error in \Cref{prop:approx_error} is negligible.

\begin{assumption}[Negligible Approximation Error of Best Predictor]\label{assump:small_approx_error}
   Assume there exists a best predictor $p$ such that for a chosen $M$ value, the uniform approximation error in \Cref{prop:approx_error} is negligible for a predictor $p$, \ie $y_{t+1+j} = \hat{h}_j^p\left(y_{t-\tau:t};\alpha^{j,p}_{\star}\right)$, where $\alpha^{j,p}_{\star} \in \calD_j$, $j\in\{0,\dots,N-1\}$, and $t\in\{1,\dots,T\}$.
\end{assumption}

\begin{theorem}[No-Regret]\label{theorem:regret}
Consider Assumptions \ref{assump:bounded_state}-\ref{assump:small_approx_error},  $\eta=\calO\left({1}/{\sqrt{T}}\right)$, $\lambda=\calO\left({1}/{\sqrt{T}}\right)$, and $\beta=\calO\left({1}/{T}\right)$. Then, the proposed method achieves
\begin{equation}
    \mathbb{E}\left[\DReg\right] \leq \calO\left(T^{\frac{3}{4}}\right).
    \label{eq:theorem_regret}
\end{equation}
\end{theorem}

\Cref{theorem:regret} implies the convergence to the clairvoyant \MPC policy $\pi^\star$ that knows the future target states, since the average regret is $\calO(T^{-\frac{1}{4}})$ and $\lim_{T\rightarrow\infty}\DReg / T \rightarrow 0$. 

For $\eta=\mathcal{O}(1/\sqrt{T})$, $\lambda=\mathcal{O}(1/\sqrt{T})$, and $\beta=\mathcal{O}(1/T)$, these parameters are \textit{fixed} throughout the horizon $T$ and scale as $1/\sqrt{T}$ or $1/T$, up to constant factors. Therefore, these parameters need to be tuned in practice, as discussed in \Cref{app:impl_details}.

When we relax \Cref{assump:small_approx_error} such that the approximation error is non-zero but the target dynamics still do \underline{not} switch,  then \cref{eq:theorem_regret} generalizes as follows, where $\epsilon_t^j$ is the approximation error of the $j$-step predictor at $t$, \ie for a predictor $p$ and parameters $\alpha^{j,p}_{\star}\in\calD_j$ as in \Cref{assump:small_approx_error},
\begin{equation}
    \epsilon_t^j \triangleq y_{t+1+j}-\hat{h}_j^p\left(y_{t-\tau:t};\alpha^{j,p}_{\star}\right), \quad j\in\{0,\dots,N-1\}.
    \label{eq:approx_error}
\end{equation}
$\epsilon_t^j$ can also encapsulate measurement noise, etc.

\begin{corollary}[Regret in the Presence of Approximation Errors] \label{cor:regret_with_error}
In presence of $\epsilon_t^j$,  the proposed method achieves
\begin{equation}
  \mathbb{E}\left[\DReg\right]
  \;\le\;
  \mathcal{O}\!\left(T^{3/4}+\sqrt{\,T \sum_{t=1}^{T} \sum_{j=0}^{N-1} \lVert \epsilon_t^j\rVert^{2}}\right).
\end{equation}
\end{corollary}
The regret scales proportionally to the root-mean-square error (and if the error $\epsilon_t^j$ is zero for all $t$ and $j$, then we recover \Cref{theorem:regret}). 
Intuitively, if $\|\epsilon_t^j\|\leq \epsilon_{\text{max}}$  for all $t$ and $j$, then $\mathrm{Regret}^{D}_{T}/T=\calO(T^{-\frac{1}{4}}+\epsilon_{\text{max}})$, \ie guarantees average regret within a radius $\calO(\epsilon_{\text{max}})$ from the clairvoyant \MPC policy, as $T\rightarrow\infty$.

\begin{corollary}[Regret Guarantees under Switching Target Dynamics]\label{cor:switching}
Partition the time horizon into $\kappa+1$ consecutive intervals $I_1,\dots,I_{\kappa+1}$, such that on each interval $I_k$ there exist a predictor $p^\star_k$ and parameters $\alpha^{j}_{\star,k}\in\calD_j$, $j\in\{0,\dots,N-1\}$, and define the approximation error, for $t\in I_k$, as $\epsilon_t^j \triangleq y_{t+1+j}-\hat{h}_j^{p^\star_k}\left(y_{t-\tau:t};\alpha^{j}_{\star,k}\right)$. Consider Assumptions \ref{assump:bounded_state}-\ref{assump:perturbation}, and the same parameters as in \Cref{theorem:regret}, \ie $\eta=\calO\left({1}/{\sqrt{T}}\right)$, $\lambda=\calO\left({1}/{\sqrt{T}}\right)$, and $\beta=\calO\left({1}/{T}\right)$, which do not depend on $\kappa$. Then, the proposed method's regret from \Cref{theorem:regret} generalizes to 
\begin{equation}
    \begin{aligned}
      \mathbb{E}\left[\DReg\right]
      \;\le\; &
      \mathcal{O}\!\left(\sqrt{\kappa+1}\,T^{\frac{3}{4}}+\sqrt{\,T \sum_{t=1}^{T} \sum_{j=0}^{N-1}\lVert \epsilon_t^j\rVert^{2}}\right) \\
      & \;+\;
      \tilde{\calO}\left((\kappa+1)\sqrt{T}\right),
    \end{aligned}
    \label{eq:corollary_switching}
\end{equation}
where $\tilde{\calO}(\cdot)$ hides the logarithmic terms of $T$, \ie $\log(T)$. 
\end{corollary}
$\kappa$ captures the intuition that the more frequent the switches are, the more challenging the online learning problem is. Several fundamental limits have been proved that lower bound the regret of online learning in non-stationary environments as an increasing function of $\kappa$~\citep{lattimore2020bandit}.  
Since the parameters of the proposed method do not depend on $\kappa$, \cref{eq:corollary_switching} holds for any such partition of the time horizon, trading off the number of switches $\kappa$ against the approximation errors $\epsilon_t^j$. The average regret vanishes whenever $\kappa=o(\sqrt{T})$ and the approximation errors vanish; in this case, the term $\tilde{\calO}((\kappa+1)\sqrt{T})$ is dominated by $\calO(\sqrt{\kappa+1}\,T^{\frac{3}{4}})$. For $\kappa=0$, \Cref{cor:switching} recovers \Cref{cor:regret_with_error}.

The theoretical guarantees in \Cref{theorem:regret}, \Cref{cor:regret_with_error}, and \Cref{cor:switching} predict an upper bound on $\text{Regret}_T^D$. In the next section, we present simulation and hardware experiments in which the targets show diverse behavior, and compare the tracking performance with baseline methods. In \Cref{app:regret}, we empirically validate the theoretical results of regret under diverse target behaviors. 
Specifically, with smooth, predictable, and repetitive trajectories, we observe decreasing average regret, $\text{Regret}_T^D/T$, as predicted by \Cref{theorem:regret}. In case of non-adversarial switching or random target dynamics, we observe bounded or mildly increasing average regret, $\text{Regret}_T^D/T$ that is predicted by \Cref{cor:regret_with_error} and \Cref{cor:switching}.

\section{Experiments}\label{sec:results}

We deploy our algorithm on Crazyflie drone for target tracking scenarios of pursuit-evasion. Particularly, we consider challenging scenarios where the pursuer is required to track a target whose dynamics exhibit structured, random, adversarial, and even switching behaviors. 
We detail experiment setup in \Cref{subsec:setup} and present simulation and hardware results in \Cref{subsec:sim_results} and \Cref{subsec:hardware}.
We analyze the individual components of
the proposed method, \ISO and \AS, in \Cref{subsec:indep_comp}.

\begin{remark}
The runtime of our implementation scales linearly with the number $P$ of predictors, since we do not parallelize computations.  For $P=4$, $\tau=10$, and $M=50$, the runtime for online update of \ISO and \AS is $9.0\pm3.2\;ms$ on a laptop with Intel i9 13th-gen processor.
\end{remark}

\subsection{Experiment setup}\label{subsec:setup}

\myParagraph{Dynamics} 
The dynamics of a Crazyflie quadrotor are~\citep{llanes2024crazysim}:
\begin{equation}
    \dot{\boldsymbol{p}} = \boldsymbol{v}, \;
    m \dot{\boldsymbol{v}} = m \boldsymbol{g} + \boldsymbol{f},  \;
    \dot{\boldsymbol{\psi}} = \frac{\boldsymbol{\psi}_c - \boldsymbol{\psi}}{t_c},
\end{equation}
where $\boldsymbol{p} \in \mathbb{R}^{3}$ and $\boldsymbol{v} \in \mathbb{R}^{3}$ are position and velocity in the inertial frame, $\boldsymbol{\psi}$ is the Euler angle, $m$ is the quadrotor mass, $\boldsymbol{g}$ is the gravity vector, $\boldsymbol{f} =  \boldsymbol{R}\left[0\ 0\ f_t\right]^\top\in \mathbb{R}^3$, $\boldsymbol{R}\in SO(3)$ is the rotation matrix, $f_t$ is the thrust from the four rotors along the $z-$axis of the body frame, and $\boldsymbol{\psi}_c \in \mathbb{R}^3$ is the commanded attitudes, and $t_c$ is the time constant of the attitude dynamics.  

\myParagraph{Benchmark Algorithms} 
We compare our algorithm to the \PLOT~\citep{tsiamis2024predictive}, the \MODL~\citep{valkanas2025modl}, and the \SSI adapted from \cite{zhou2025simultaneous}. 
All the algorithms learn the target dynamics fully online.
\PLOT predicts target positions with multiple linear motion models.
\MODL uses three predictors: online linear regression, a multi-layer perceptron, and a set-learning residual neural network in tandem. 
For  \SSI, we use the system identification model based on \RFF to recursively predict the multi-step trajectory for the target. 
Contrary to the multi-step prediction model in \ISO and \PLOT, both \SSI and \MODL use a single-step learned dynamics model to recursively predict the multi-step trajectory for the target. 
For all methods, the predicted target trajectory is then used as a reference for the \MPC controller for the pursuer. 

We report the results for \PLOT with tuning parameters $\gamma=0.5$ and $\gamma=0.9$ and for \SSI with std. dev. of Gaussian kernel $\sigma=0.1$ and $\sigma=1.0$. 
We evaluate all algorithms via tracking error and prediction accuracy.
Tracking error quantifies the distance between the pursuer and the target at each time step. We report the mean error over the duration of the experiment and the final error.
Prediction accuracy is a retrospective metric that quantifies the error of the \ISO predictions, given the observed target states. Particularly, at each time step across a past \MPC horizon, the error is calculated given the past stored predictions of the target state and the observed target state. 

\myParagraph{Target} 
The target drone moves using an \MPC controller, whose predictions (intended positions) are accessible to a clairvoyant controller \OPT. For a non-adversarial target, this information can be viewed as the ground truth target position. For an adversarial target, this intention is subject to change as it may adapt to the pursuer's tracking performance. This determines the best possible tracking performance for any kind of target, keeping the other \MPC parameters same. 

Our algorithm tracks all target types with its hyperparameters fixed across all experiments.  
In all algorithms, the \MPC parameters, such as prediction horizon and cost functions, are the same.
Additional details about implementation and parameters used can be found in \Cref{app:impl_details}. 

\subsection{Simulation Experiments}\label{subsec:sim_results}

\begin{table*}[t]
\centering
\caption{Mean tracking error (m) over 10 runs per method and target. In brackets: how higher ($+$) or lower ($-$) the error is compared to {Ours}. The last row averages the mean errors over the six target types. The lowest error excluding \OPT is shown in \textbf{bold}, the second lowest is \underline{underlined}. $^{\S}$On Sinusoidal trajectory, the \PLOT columns use $\gamma{=}0.95$ and $\gamma{=}1.0$, respectively, since for $\gamma \leq0.9$ the prediction becomes massively incorrect and unstable. \OPT has knowledge of the target's future position obtained from the target's \MPC, thus achieving the best performance.}
\label{tab:tracking_error_nostd}
\resizebox{\textwidth}{!}{%
\begin{tabular}{lccccccc}
\toprule
Target & \PLOT ($\gamma{=}0.5$) & \PLOT ($\gamma{=}0.9$) & \RFF ($\sigma{=}0.1$) & \RFF ($\sigma{=}1.0$) & \MODL & Optimal$^{*}$ & {Ours} \\
\midrule
Lemniscate & $0.139$\,{\scriptsize(+42\%)} & $0.484$\,{\scriptsize(+395\%)} & $0.215$\,{\scriptsize(+120\%)} & \underline{$0.114$}\,{\scriptsize(+17\%)} & $0.223$\,{\scriptsize(+128\%)} & $0.065$\,{\scriptsize($-$33\%)} & $\mathbf{0.098}$ \\
Lissajous & \underline{$0.216$}\,{\scriptsize(+6\%)} & $0.675$\,{\scriptsize(+231\%)} & $0.454$\,{\scriptsize(+122\%)} & $0.265$\,{\scriptsize(+30\%)} & $0.383$\,{\scriptsize(+88\%)} & $0.109$\,{\scriptsize($-$46\%)} & $\mathbf{0.204}$ \\
Sinusoidal & $0.735^{\S}$\,{\scriptsize(+176\%)} & $0.560^{\S}$\,{\scriptsize(+111\%)} & \underline{$0.302$}\,{\scriptsize(+13\%)} & $0.547$\,{\scriptsize(+106\%)} & $0.391$\,{\scriptsize(+47\%)} & $0.108$\,{\scriptsize($-$59\%)} & $\mathbf{0.266}$ \\
Star & $0.167$\,{\scriptsize(+59\%)} & $0.229$\,{\scriptsize(+118\%)} & \underline{$0.130$}\,{\scriptsize(+24\%)} & $0.165$\,{\scriptsize(+58\%)} & $0.200$\,{\scriptsize(+91\%)} & $0.045$\,{\scriptsize($-$57\%)} & $\mathbf{0.105}$ \\
Random Walk & $0.364$\,{\scriptsize(+124\%)} & $0.383$\,{\scriptsize(+136\%)} & \underline{$0.171$}\,{\scriptsize(+5\%)} & $0.311$\,{\scriptsize(+92\%)} & $0.292$\,{\scriptsize(+80\%)} & $0.067$\,{\scriptsize($-$59\%)} & $\mathbf{0.162}$ \\
Random Adversarial & $0.475$\,{\scriptsize(+43\%)} & $0.531$\,{\scriptsize(+59\%)} & \underline{$0.344$}\,{\scriptsize(+3\%)} & $0.464$\,{\scriptsize(+39\%)} & $0.456$\,{\scriptsize(+37\%)} & $0.303$\,{\scriptsize($-$9\%)} & $\mathbf{0.333}$ \\
\midrule
Average & $0.349$\,{\scriptsize(+79\%)} & $0.477$\,{\scriptsize(+145\%)} & \underline{$0.269$}\,{\scriptsize(+38\%)} & $0.311$\,{\scriptsize(+60\%)} & $0.324$\,{\scriptsize(+67\%)} & $0.116$\,{\scriptsize($-$40\%)} & $\mathbf{0.195}$ \\
\bottomrule
\end{tabular}}
\label{table:sim}
\end{table*}

\begin{figure*}[t]
    \centering
    \includegraphics[width=\textwidth]{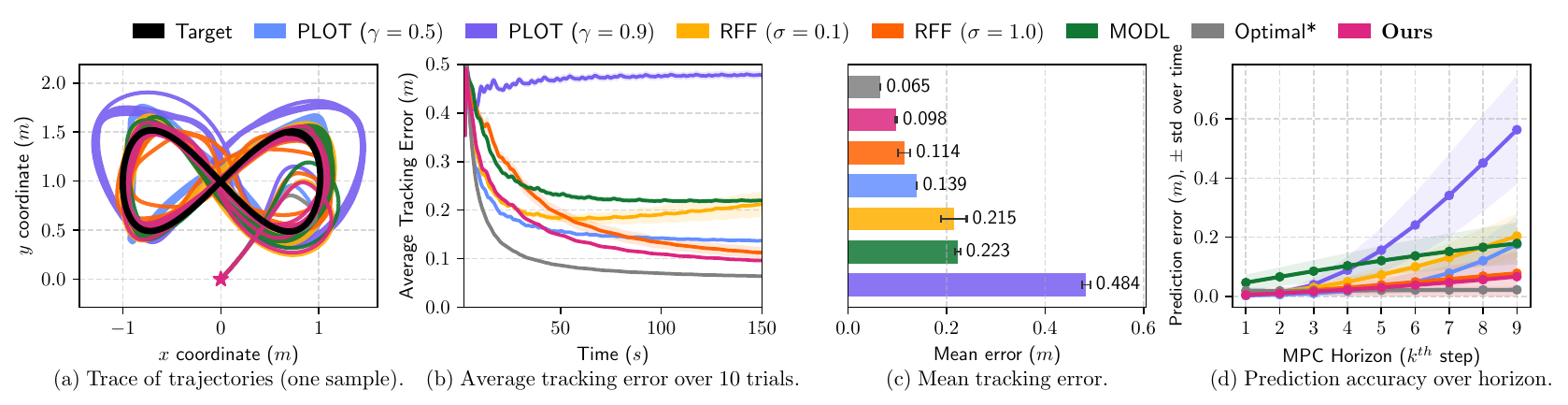}
    \refstepcounter{figure}
    \refstepcounter{subfigure}
    \label{fig:lemniscate_a}
    \refstepcounter{subfigure}
    \label{fig:lemniscate_b}
    \refstepcounter{subfigure}
    \label{fig:lemniscate_c}
    \refstepcounter{subfigure}
    \label{fig:lemniscate_d}
    \addtocounter{figure}{-1}
    \vspace{-5mm}    
    \caption{\textbf{Lemniscate target trajectory.} (a) Trace of trajectories shows that our algorithm achieves qualitatively better tracking performance compared to baselines. (b) The average tracking error plot shows an asymptotically decreasing curve of our algorithm that is converging towards the \OPT. (c) The mean tracking error ($m$) shows that our algorithm achieves the lowest tracking error.  (d) The retrospective prediction error also shows similar or better performance to the \PLOT, \SSI, and \MODL.}
    \label{fig:lemniscate}
\end{figure*}

\begin{figure*}[t]
    \centering
    \includegraphics[width=\textwidth]{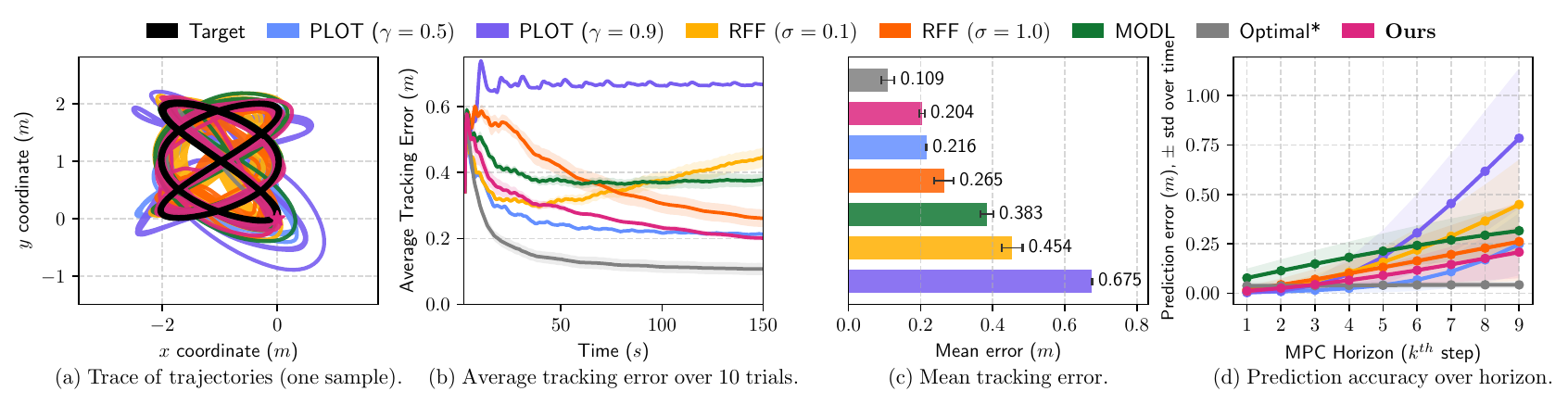}
    \refstepcounter{figure}
    \refstepcounter{subfigure}
    \label{fig:lissajous_a}
    \refstepcounter{subfigure}
    \label{fig:lissajous_b}
    \refstepcounter{subfigure}
    \label{fig:lissajous_c}
    \refstepcounter{subfigure}
    \label{fig:lissajous_d}
    \addtocounter{figure}{-1}
    \vspace{-5mm}
    \caption{\textbf{Lissajous target trajectory.} (a) Trace of trajectories. (b) \& (c) The tracking error ($m$) shows that \PLOT~($\gamma = 0.5$) has comparable tracking performance to our algorithm, and our algorithm is substantially better than other baselines. (d) Retrospective prediction error shows a similar trend to that in (b) \& (c).}
    \label{fig:lissajous}
\end{figure*}

\begin{figure*}[t]
    \centering
    \includegraphics[width=\textwidth]{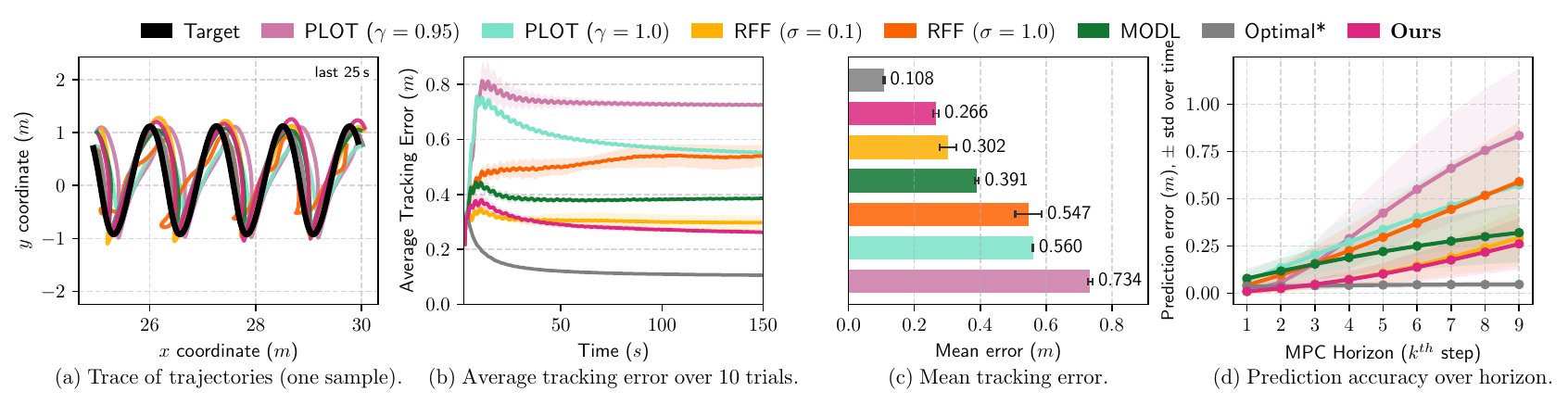}
    \refstepcounter{figure}
    \refstepcounter{subfigure}
    \label{fig:sinusoidal_a}
    \refstepcounter{subfigure}
    \label{fig:sinusoidal_b}
    \refstepcounter{subfigure}
    \label{fig:sinusoidal_c}
    \refstepcounter{subfigure}
    \label{fig:sinusoidal_d}
    \addtocounter{figure}{-1}
    \vspace{-5mm}
    \caption{\textbf{Sinusoidal target trajectory.} (a) Trace of trajectories. (b) \& (c) The tracking error ($m$) shows superior performance by our algorithm. We consider high values of $\gamma \in \{0.95, 1.0\}$ for \PLOT since smaller values lead to a crash due to poor predictions. (d) Retrospective prediction error shows a similar trend to that in (b) \& (c).}
    \label{fig:sinusoidal}
\end{figure*}

\begin{figure*}[t]
    \centering
    \includegraphics[width=\textwidth]{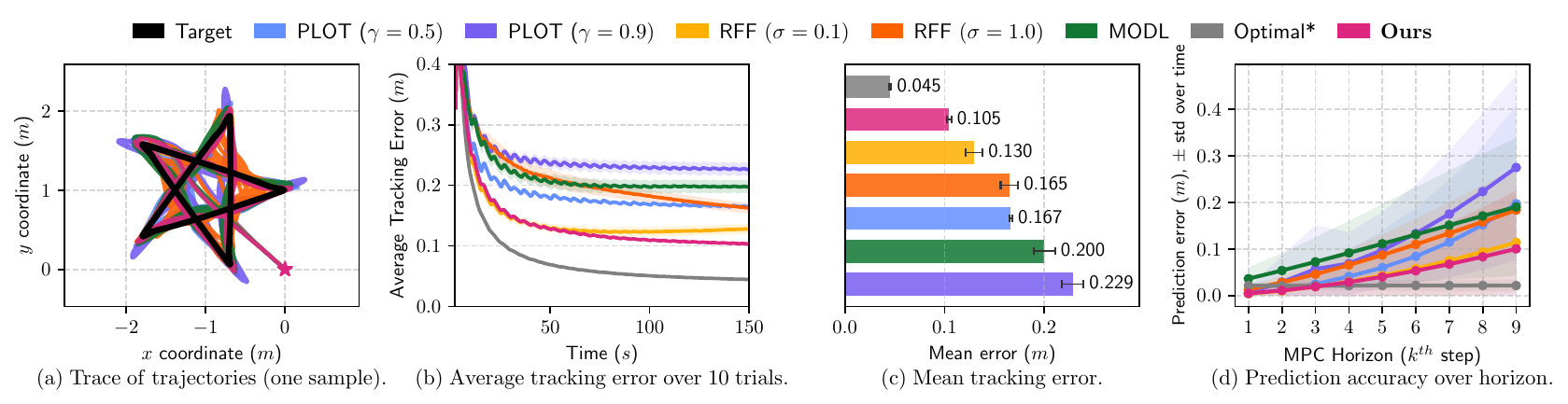}
    \refstepcounter{figure}
    \refstepcounter{subfigure}
    \label{fig:star_a}
    \refstepcounter{subfigure}
    \label{fig:star_b}
    \refstepcounter{subfigure}
    \label{fig:star_c}
    \refstepcounter{subfigure}
    \label{fig:star_d}
    \addtocounter{figure}{-1}
    \vspace{-5mm}
    \caption{\textbf{Star target trajectory.} (a) Trace of trajectories. (b) \& (c) Our algorithm has substantially better tracking performance than other baselines. (d) Retrospective prediction error shows a similar trend to that in (b) \& (c).}
    \label{fig:star}
\end{figure*}

\begin{figure*}[t]
    \centering
    \includegraphics[width=\textwidth]{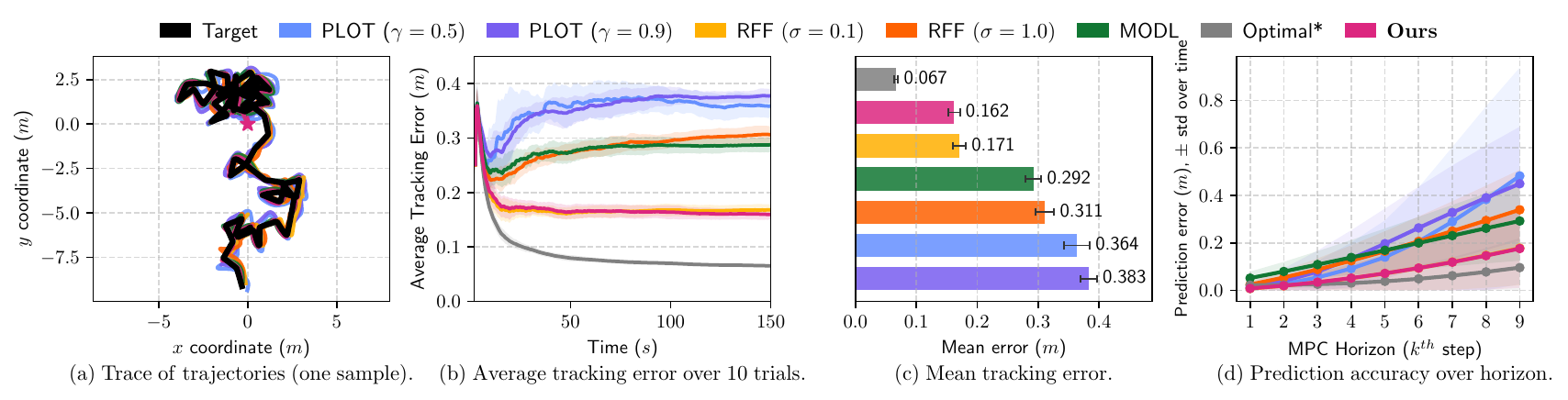}
    \refstepcounter{figure}
    \refstepcounter{subfigure}
    \label{fig:random_walk_a}
    \refstepcounter{subfigure}
    \label{fig:random_walk_b}
    \refstepcounter{subfigure}
    \label{fig:random_walk_c}
    \refstepcounter{subfigure}
    \label{fig:random_walk_d}
    \addtocounter{figure}{-1}
    \vspace{-5mm}
    \caption{\textbf{Random walk target.}  (a) Trace of trajectories. 
    (b) \& (c) The tracking error ($m$) shows that \RFF~($\sigma = 0.1$) has comparable tracking performance to our algorithm, and our algorithm is substantially better than other baselines. 
    (d) Retrospective prediction error shows a similar trend to that in (b) \& (c).}
    \label{fig:random_walk}
\end{figure*}

\begin{figure*}[t]
    \centering
    \includegraphics[width=\textwidth]{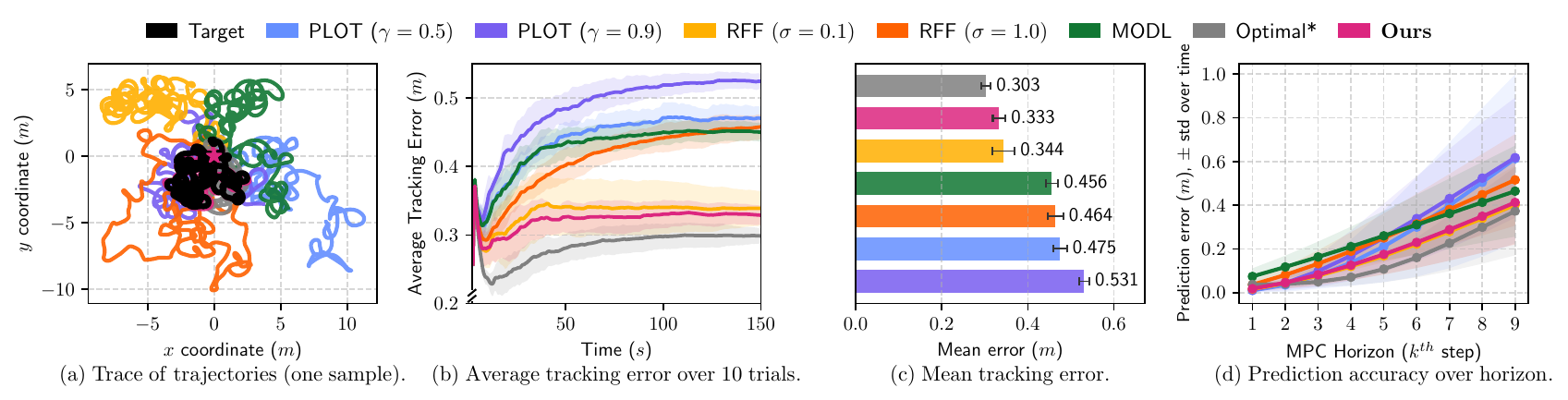}
    \refstepcounter{figure}
    \refstepcounter{subfigure}
    \label{fig:adversarial_a}
    \refstepcounter{subfigure}
    \label{fig:adversarial_b}
    \refstepcounter{subfigure}
    \label{fig:adversarial_c}
    \refstepcounter{subfigure}
    \label{fig:adversarial_d}
    \addtocounter{figure}{-1}
    \vspace{-5mm}
    \caption{\textbf{Randomized adversarial target.} (a) Trace of trajectories. In this case, the target trajectories for each algorithm are different; hence we only plot the target trajectory corresponding to our algorithm. (b) \& (c) The tracking error ($m$) shows that \RFF~($\sigma = 0.1$) has comparable tracking performance to our algorithm, and our algorithm is substantially better than other baselines. 
    (d) Retrospective prediction error shows a similar trend to that in (b) \& (c).}
    \label{fig:adversarial}
\end{figure*}

We consider six scenarios: (a) lemniscate trajectory, (b) Lissajous trajectory, (c) sinusoidal trajectory, (d) star trajectory (e) random walk, and (f) randomized adversarial. 
We present the results averaged over 10 seed-controlled runs for all the target types and baselines. \Cref{tab:tracking_error_nostd} summarizes the main quantitative results in terms of average tracking error, while \Cref{fig:lemniscate,fig:lissajous,fig:sinusoidal,fig:star,fig:random_walk,fig:adversarial} depict the trajectories, tracking errors evolved over time, and prediction accuracy.
In more detail:

\paragraph{Lemniscate.} The target moves through a closed-loop, two-dimensional lemniscate trajectory as shown in \Cref{fig:lemniscate_a}. Particularly, it is given by $p_x(t) = \sin(0.5 t),\; p_y(t) = 1.0 + \sin(0.5 t) \cos(0.5 t)$.
Being a closed-loop deterministic trajectory, this is the simplest tested case for target tracking. As seen in \Cref{tab:tracking_error_nostd}, \PLOT with $\gamma=\{0.5,0.9\}$ show large differences in performance, while the \SSI algorithm has lower variation for $\sigma=\{0.1,1.0\}$. With limited onboard data, \MODL converges slowly to give a near-constant tracking error. Online deep learning is typically slow and more data-hungry due to online backpropagation step. 
\Cref{table:sim} shows that our method achieves the lowest tracking error of $0.098$m, while the next best method, \RFF ($\sigma{=}1.0$), has $17\%$ higher average tracking error. While \PLOT ($\gamma{=}0.5$) shows faster convergence (\Cref{fig:lemniscate_b}) in first few seconds due to second-order recursive least-squares, our method shows better overall tracking and prediction performance.

\paragraph{Lissajous.} The target moves through a two-dimensional $3:2$ ratio Lissajous curve trajectory (\Cref{fig:lissajous_a}) given by $p_x(t) = \sin\!\big(0.9t + \pi/2\big),\; p_y(t) = \sin\!\big(0.6t\big)$.
Compared to the lemniscate, this target moves aggressively, with more than $1.5\times$ both speed and acceleration, in a slightly more complex closed-loop trajectory. \Cref{table:sim} shows that our method achieves the lowest tracking error, while \PLOT ($\gamma=0.5$) has similar performance with $6\%$ more average tracking error across all trials. Previously-best baseline, \RFF ($\sigma{=}1.0$), now has $30\%$ larger tracking error than ours. From \Cref{fig:lissajous_b}, we observe a similar trend as before where \PLOT has faster convergence but our method surpasses the performance eventually. As we see next, while \PLOT maintains strong performance in smooth closed-loop trajectories, it degrades sharply in open-loop or non-smooth trajectories.

\paragraph{Sinusoidal.} The target moves through an open-loop sinusoidal trajectory as shown in \Cref{fig:sinusoidal}. This trajectory is predictable, but not a closed-loop trajectory, hence more challenging to learn than the previous, since the target does not return to the same position. 
The two-dimensional target trajectory is given by $p_x(t) = 0.2 t$, $p_y(t) = -\sin(t)$.
For the \PLOT, we report results for $\gamma \in \{0.95, 1.0\}$: we observed the predictions for $\gamma\leq0.9$ to be massively incorrect and unstable. 
Our method maintains best tracking performance, with \RFF ($\sigma=0.1$) being the second with $13\%$ higher tracking error.
In contrast to the lemniscate and Lissajous trajectory, a higher value of $\gamma$ for \PLOT and lower $\sigma$ for \RFF have better performance here. This suggests these methods are unable to adapt to the unknown target dynamics for a fixed set of hyperparameters.

\paragraph{Star.}
The target moves through a closed-loop, two-dimensional star-shaped trajectory, as shown in \Cref{fig:star}, constructed using straight lines and sharp turns. \Cref{tab:tracking_error_nostd} shows that our method achieves the lowest tracking error, with the next-best \RFF ($\sigma=0.1$) having a $24\%$ higher tracking error. The best \PLOT ($\gamma=0.5$) has over $59\%$ larger tracking error, highlighting its limitations in generalizing across different target dynamics. 

\paragraph{Random Walk.} The target performs a random walk, starting at the origin, moving in a random direction with a constant speed of $0.5$ m/s for a duration of $1$s, and then selecting a new direction randomly from a uniform distribution, depicted in \Cref{fig:random_walk_a}.
\Cref{tab:tracking_error_nostd} shows that our method maintains lowest tracking error, while \RFF ($\sigma=0.1$) is a close second with $5\%$ higher error. \PLOT and \MODL have considerably worse tracking in comparison to ours.

\paragraph{Randomized Adversarial.} The target follows a trajectory with random direction changes as a function of the pursuer's position. Specifically, the target policy is to always move in the half-space opposite to the pursuer, with a speed that is inversely proportional to their distance. In addition, the direction in the half-space (say angle $\theta_{\text{target}}\in[\pi/2, -\pi/2]$) changes randomly every 1$s$. Rigorously, the  trajectory is described by $\|\boldsymbol{v}_{\text{target}}\|_2=\;\exp\left(-0.75 \|\boldsymbol{p}_{\text{target}} - \boldsymbol{p}_{\text{pursuer}}\|_2\right)$, $\theta_{\text{target}} \sim \mathcal{N}(-\pi/2, \pi/2) \text{ every 1}s$,
where $\theta_{\text{target}}$ is the angle between target velocity $\boldsymbol{v}_{\text{target}}$ and outward vector $\boldsymbol{p}_{\text{target}} - \boldsymbol{p}_{\text{pursuer}}$. $\boldsymbol{p}_{\text{target}}$ and $\boldsymbol{p}_{\text{pursuer}}$ are the current target and pursuer positions, respectively. Even for a constant value of $\theta_{\text{target}}$, the target is adversarial and does not move in a straight line, since the half-plane is dynamic and opposite to the pursuer's position.  

Since the adversarial target motion depends on the pursuer, our theoretical analysis on regret does not hold in this scenario. However, we observe empirically that our method has the closest performance to \OPT.
\SSI~($\sigma = 0.1$) remains similar to our algorithm in this case, while all other baselines are at least $37\%$ worse than ours.

\paragraph{Prediction Accuracy.} 

For each of the experiments in \Cref{fig:lemniscate} to \ref{fig:adversarial}, we plot the prediction accuracy (\Cref{fig:lemniscate_d} -- \ref{fig:adversarial_d}) of each method over the \MPC time horizon. We observe that our method has consistently lower prediction error than other methods, a trend similar to the tracking error (\Cref{fig:lemniscate_b} -- \ref{fig:adversarial_b} \& \Cref{fig:lemniscate_c} -- \ref{fig:adversarial_c}). The tracking error is comparable to the best-suited \SSI method in some cases, but 
We observe that \SSI has higher jitter due to the compounding effect of learning error of the single learning function. This is also reflected in generally higher standard deviation in tracking errors for \RFF methods (\Cref{fig:lemniscate_c} -- \ref{fig:adversarial_c}) compared to our method, even in case of non-random target.  
Overall, averaged over different target types in \Cref{tab:tracking_error_nostd}, our method has the best performance among all the algorithms. In particular, \RFF ($\sigma=0.1$) has comparable performance (within $3-13\%$) in randomized, sinusoidal, and star trajectories, but more than $120\%$ worse tracking error for smooth trajectories of lemniscate and Lissajous. Across all considered target behaviors and trials, the next-best baseline has $38\%$ higher tracking errors than our method, demonstrating strong learning and adaptation capabilities of our method.

\begin{figure}[t]
    \centering
    \includegraphics[width=\linewidth]{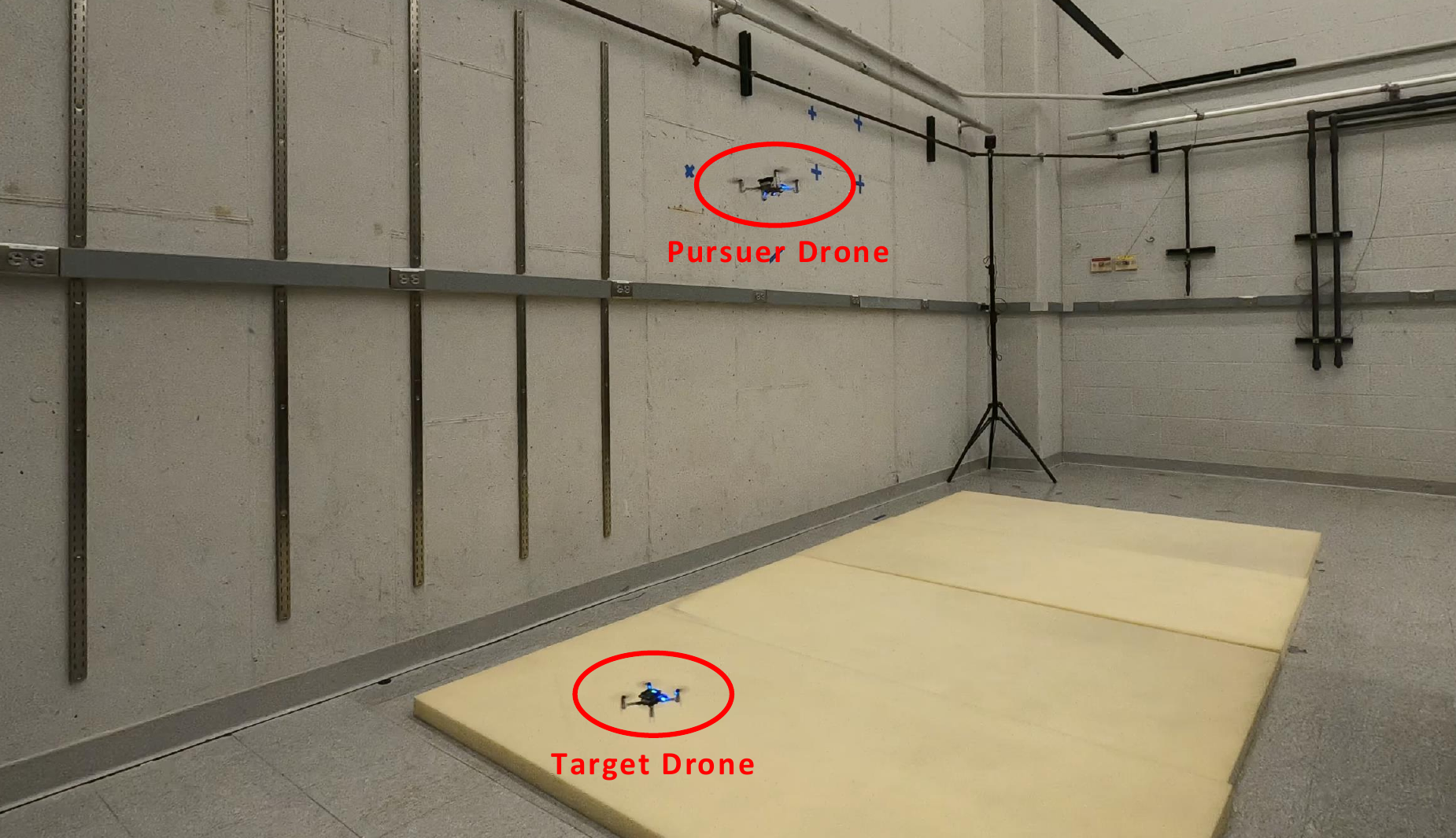}
    \caption{\textbf{Setup for hardware experiments.} Two physical Crazyflie drones are used, one as a pursuer and one as a target. To avoid collisions, a vertical offset between the pursuer and target drone is maintained. Due to the vertical offset, the target drone is also affected by the downwash effect of the pursuer drone, especially when the pursuer drone is able to closely track the target drone. Such downwash effects add complexity and randomness to the target drone behavior. }
    \label{fig:hw_snapshot}
\end{figure}

\subsection{Hardware Experiments}\label{subsec:hardware}

We validate our algorithm on the Crazyflie hardware and consider a target changing its motion online. 
The hardware setup is shown in \Cref{fig:hw_snapshot}, where we use two physical Crazyflie drones.
We consider diverse target types following hybrid dynamics, to demonstrate the effectiveness of our algorithm. In particular, we consider (a) a target that starts with a smooth and periodic lemniscate trajectory, then changes to a randomized adversarial trajectory, and again switches back to a lemniscate trajectory~(\Cref{fig:hardware_a}), (b) a target that first follows a star-shaped trajectory with sharp corners and then switches to a smooth Lissajous curve trajectory~(\Cref{fig:hardware_b}), and (c) a target following the shape of alphabetical letters "UM"~(\Cref{fig:hardware_c}). To avoid collisions, we maintain a vertical offset between the pursuer and target drone, while the target dynamics are considered in the $X-Y$ plane only.  Except for the unknown and non-stationary trajectory, the target drone is also affected by the downwash effect of the pursuer drone, especially when the pursuer drone is able to closely track the target drone. Such downwash effects add complexity and randomness to the target drone behavior.

\Cref{table:real} and \Cref{fig:hardware} show that our algorithm maintains the lowest average tracking error throughout all three experiments, even when the target changes behavior online. Specifically, our algorithm on average improves tracking performance by at least $38\%$ compared to baselines. Any instance of \PLOT or \SSI has good tracking performance only during a certain type of target type, and has deteriorating performance when the target switches behavior. For example, \PLOT with $\gamma=0.5$ has good tracking performance in predictable smooth trajectories, \ie lemniscate and lissajous, but suffers larger tracking errors in random or non-smooth trajectories. Similarly, \SSI with $\sigma=0.1$ has good performance when the target trajectory consists of constant-velocity segments in star-shape or “UM” letters, but has worse performance in other kinds of trajectories. 
This demonstrates the adaptability of our method to changing target types in contrast to the compared algorithms. Overall, we found that \MODL has poor adaptability for out-of-distribution datasets, showcasing the inefficiency of the deep learning models in terms of training data. 

\begin{table*}[t]
\centering
\caption{Mean tracking error (m) per method in the hardware experiments (one flight per method; scored windows: (a) Lemniscate$\rightarrow$Adversarial$\rightarrow$Lemniscate (260\,s), (b) Star$\rightarrow$Lissajous (200\,s), and, (c) UM letters (100\,s). In brackets: how much higher the error, in percent, of that method compared to Ours. The last row averages the mean errors over the three experiments. The lowest error excluding \OPT is shown in \textbf{bold}; the second lowest is \underline{underlined}. \OPT has knowledge of the target's future position obtained from the target's \MPC,  thus achieving the best performance.}
\label{tab:hardware_error}
\resizebox{\textwidth}{!}{%
\begin{tabular}{lccccccc}
\toprule
Target & \PLOT ($\gamma{=}0.5$) & \PLOT ($\gamma{=}0.9$) & \RFF ($\sigma{=}0.1$) & \RFF ($\sigma{=}1.0$) & \MODL & Optimal$^{*}$ & Ours \\
\midrule
Lemni.$\rightarrow$Adv.$\rightarrow$Lemni. & $0.265$\,{\scriptsize(+47\%)} & $0.500$\,{\scriptsize(+177\%)} & $0.273$\,{\scriptsize(+51\%)} & \underline{$0.244$}\,{\scriptsize(+35\%)} & $0.285$\,{\scriptsize(+58\%)} & $0.126$\,{\scriptsize($-$30\%)} & $\mathbf{0.181}$ \\
Star$\rightarrow$Lissajous & \underline{$0.161$}\,{\scriptsize(+4\%)} & $0.353$\,{\scriptsize(+127\%)} & $0.226$\,{\scriptsize(+45\%)} & $0.215$\,{\scriptsize(+38\%)} & $0.233$\,{\scriptsize(+50\%)} & $0.068$\,{\scriptsize($-$56\%)} & $\mathbf{0.155}$ \\
UM letters & $0.293$\,{\scriptsize(+72\%)} & $0.386$\,{\scriptsize(+127\%)} & \underline{$0.201$}\,{\scriptsize(+18\%)} & $0.264$\,{\scriptsize(+55\%)} & $0.248$\,{\scriptsize(+45\%)} & $0.078$\,{\scriptsize($-$54\%)} & $\mathbf{0.170}$ \\
\midrule
Average & $0.240$\,{\scriptsize(+42\%)} & $0.413$\,{\scriptsize(+145\%)} & \underline{$0.233$}\,{\scriptsize(+38\%)} & $0.241$\,{\scriptsize(+43\%)} & $0.255$\,{\scriptsize(+51\%)} & $0.091$\,{\scriptsize($-$46\%)} & $\mathbf{0.169}$ \\
\bottomrule
\end{tabular}}
\label{table:real}
\end{table*}

\begin{figure*}[t]
    \centering
\includegraphics[width=1.0\linewidth]{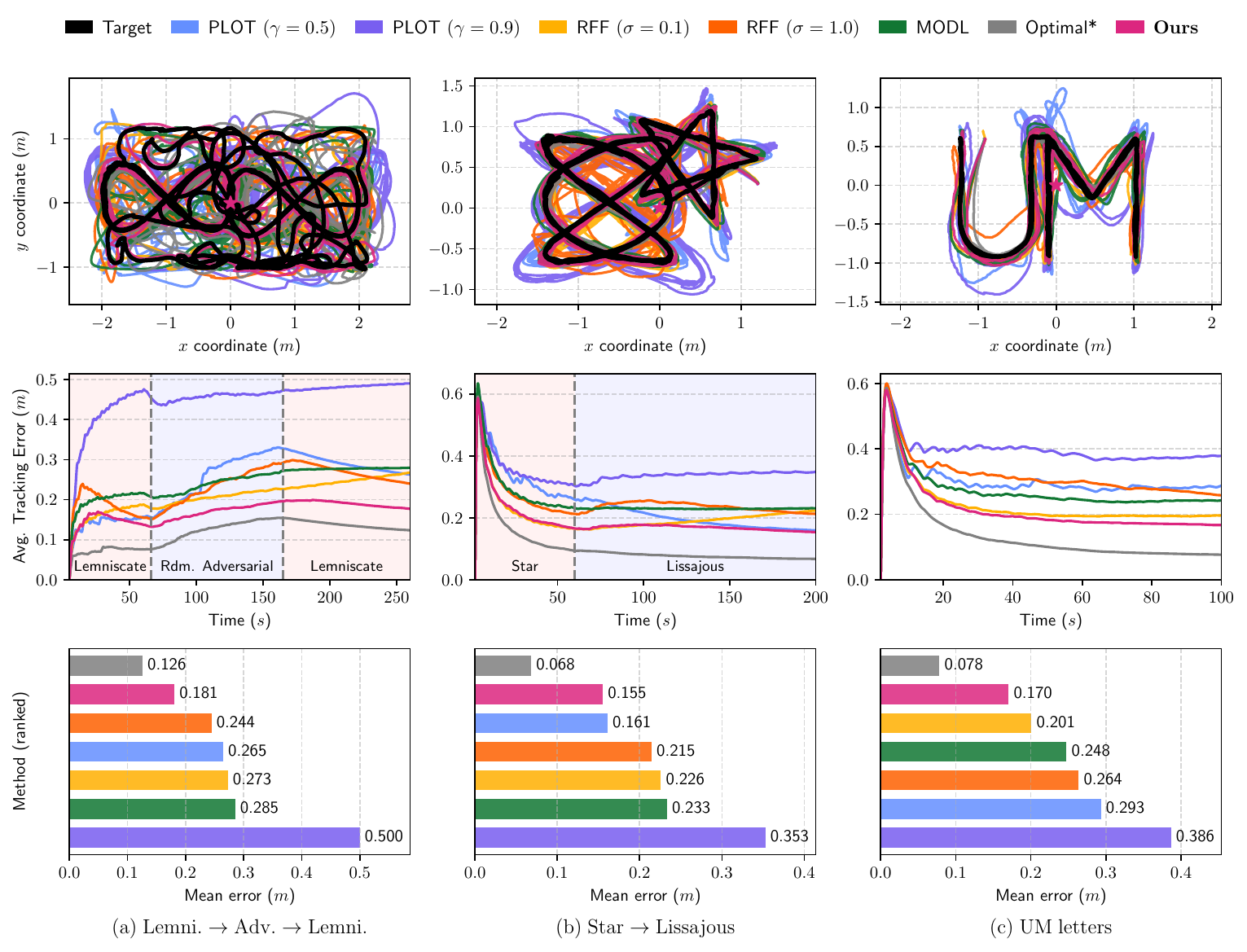}
    \refstepcounter{figure}
    \refstepcounter{subfigure}
    \label{fig:hardware_a}
    \refstepcounter{subfigure}
    \label{fig:hardware_b}
    \refstepcounter{subfigure}
    \label{fig:hardware_c}
    \addtocounter{figure}{-1}
    \vspace{-5mm}
    \caption{\textbf{Hardware experiment with 2 Crazyflie drones.}
    Three different target types are considered: (a) Hybrid target that switches between a lemniscate trajectory and randomized adversarial behavior, (b) Hybrid target that switches from a star-shaped trajectory (straight lines and sharp corners) to a smooth Lissajous curve trajectory, and (c) Target following alphabetical letters 'UM'. Our method consistently maintains better tracking performance in hardware tests than other algorithms across different target types. To avoid collision, we maintain a vertical offset between two drones and restrict their motion in the horizontal plane. } 
    \label{fig:hardware}
\end{figure*}

\subsection{Analysis of Individual Modules}\label{subsec:indep_comp}
We analyze the individual components of the proposed method. Particularly, we isolate the individual modules, \ISO and \AS, to study the effects of various parameters and conditions that can vary for different target trajectory types. We consider the following two target types:  lemniscate, and randomized adversarial. The purpose of this analysis is to show the specific challenges both modules address. The \ISO module contains an independent set of predictors that have fixed hyperparameters. We study the effect of these parameters on the performance of a single predictor using an ablation study in \Cref{subsec:ablation_study}. In this case, we remove the \AS module to analyze the tracking performance of single predictors with varying parameters. Next, for a selected set of predictors, we deploy our algorithm with the \AS module to show how it converges to the best predictor (or expert) in \Cref{subsec:hedge}. This also allows for a comparative study of the performance of our final algorithm with individual experts. 

\begin{figure}[t]
    \centering
    \includegraphics[width=\linewidth]{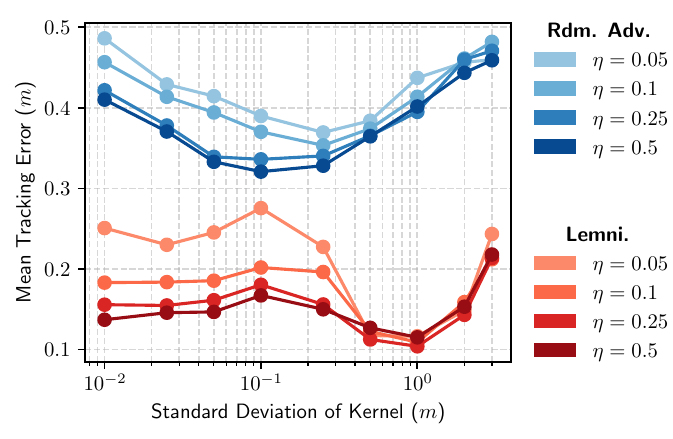}
    \caption{\textbf{
    Effect of standard deviation ${\sigma}$ and learning rate ${\eta}$.} Mean tracking error vs. standard deviation of the Gaussian kernel ${\sigma}$ for different target types and learning rates ${\eta}$. The red line is for the lemniscate trajectory target, while the blue line is for the randomized adversarial target. The optimal value of ${\sigma}$ varies significantly for different target types. Increasing ${\eta}$ generally improves the performance, but the gain is marginal or reversed after a certain value.}
    \label{fig:ablation_std}
\end{figure}

\begin{figure}[t]
    \centering
    \includegraphics[width=\linewidth]{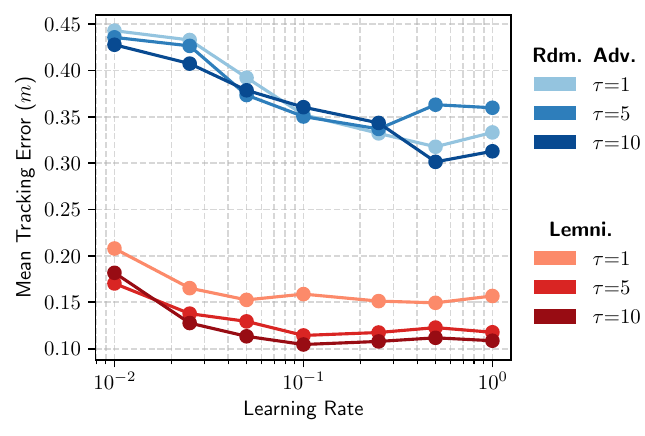}
    \caption{\textbf{Effect of learning rate $\eta$ and memory horizon of the target position $\tau$.} Mean tracking error vs. learning rate $\eta$ for different target types and memory horizon $\tau$ for the target position. Increasing learning rate generally improves the performance. Expectedly, larger memory horizon $\tau$ leads to better prediction and hence, tracking performance.}
    \label{fig:ablation_lr}
\end{figure}

\subsubsection{Ablation Study (\ISO focus)}\label{subsec:ablation_study}
For a single predictor in the \ISO module, we next observe that the hyperparameters, namely, learning rate $\eta$, standard deviation of the kernel $\sigma$, and memory horizon of target position $m_h$,  affect the tracking as well as  the prediction performance of the base predictor. Moreover, the target type and policy also affects the performance. Thus, adaptation is needed, as we introduced in this paper via \AS.  

We run each simulation for 90$s$ and compute the mean tracking error over the duration for both target types considered. First, the value of $\sigma$ of the single predictor is varied logarithmically from $\sigma=10^{-2}$ to $\sigma=3.0$. Each simulation is conducted with a different learning rate $\eta$. \Cref{fig:ablation_std} shows the plot for the lemniscate trajectory target (in red) and the randomized adversarial target (in blue). It is evident that the "optimal" $\sigma$ for the predictor varies significantly for different target type. While larger $\eta$ mostly increases the performance, the gain is marginal after a certain value. A fixed or tuned value of $\sigma$ is not guaranteed to achieve consistent performance across all target types, since the optimal value varies with the different target types. In \Cref{fig:ablation_lr}, we fix the respective optimal value of $\sigma$ for each target type \ie $\sigma=1.0$ for the lemniscate trajectory and $\sigma=0.2$ for the randomized adversarial target experiment. Then, we perform the same set of experiments with varying learning rate $\eta$ and the number of features or, the target position memory horizon $\tau$.
The performance typically increases with a higher learning rate and larger memory horizon. In the case of a randomized adversarial target, we observe that the memory horizon has little effect. This is expected since past position data from the target does not have a significant correlation with the target motion due to randomness.

\begin{figure}[t]
    \centering
    \includegraphics[width=\linewidth]{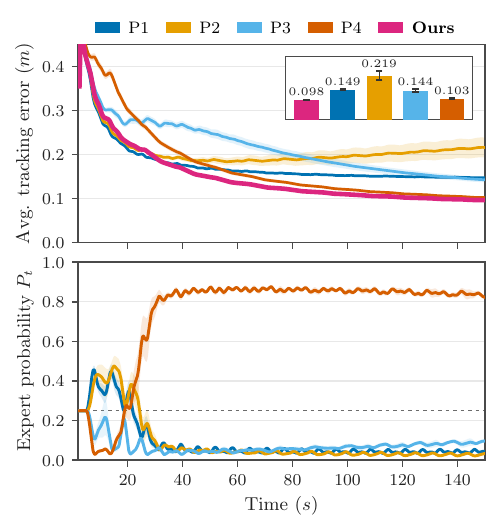}
    \refstepcounter{figure}
    \refstepcounter{subfigure}
    \label{fig:convergence_lemni_a}
    \refstepcounter{subfigure}
    \label{fig:convergence_lemni_b}
    \addtocounter{figure}{-1}
    \vspace{-5mm}
    \caption{\textbf{Lemniscate target trajectory.} (Top) The evolution of the average tracking error and the mean tracking error (inserted graph) for our algorithm is similar to the performance of the best expert. (Bottom) The probability assigned to each expert shows how the algorithm favors the \textit{agile} experts (\predA and \predB) initially, and then, as it learns, the \textit{accurate} expert \predD. This is also reflected in the individual performance of the isolated experts, shown in the inserted mean tracking error graph.}
    \label{fig:convergence_lemni}
\end{figure}

\begin{figure}[t]
    \centering
    \includegraphics[width=\linewidth]{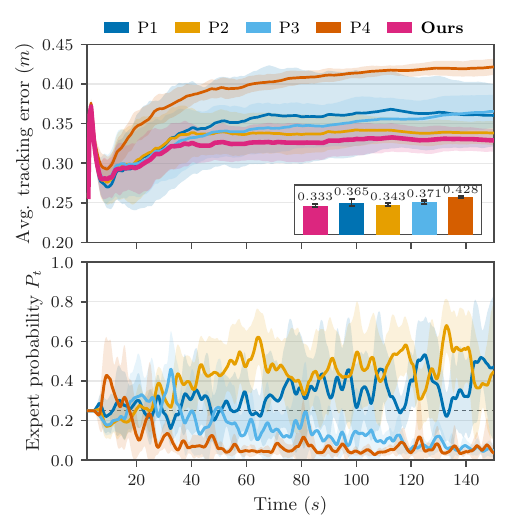}
    \refstepcounter{figure}
    \refstepcounter{subfigure}
    \label{fig:convergence_rdm_adv_a}
    \refstepcounter{subfigure}
    \label{fig:convergence_rdm_adv_b}
    \addtocounter{figure}{-1}
    \vspace{-5mm}
    \caption{\textbf{Adversarial target.} (Top) The evolution of the average tracking error and the mean tracking error (inserted graph) for our algorithm is always close to that of the best expert. (Bottom) The probability assigned to each expert shows how the algorithm favors \textit{agile} experts \predA and \predB at all times.}
    \label{fig:convergence_rdm_adv}
\end{figure}

\subsubsection{Convergence to the Best Predictor (\AS focus)}\label{subsec:hedge}

While the \AS module adapts to the target type to learn the best predictor (expert), we herein compare \AS's performance to the case where only one predictor is used at all times. In particular, we demonstrate the convergence of \AS to the best predictor for the tested target trajectories. Based on the ablation study above, we pick four predictors that differ only in the learning rate $\eta$ of the online regression and in the standard deviation $\sigma$ of the random Fourier feature frequencies: \predA ($\eta{=}0.4$, $\sigma{=}0.03$), \predB ($\eta{=}0.2$, $\sigma{=}0.1$), \predC ($\eta{=}0.1$, $\sigma{=}0.3$), and \predD ($\eta{=}0.05$, $\sigma{=}1.0$). They range from fast-adapting predictors with a high learning rate and smooth features (\predA) to slowly adapting predictors with a low learning rate and more expressive features (\predD). We compare their individual tracking performance, across the same 10 simulation experiments, with that of the final algorithm with the \AS module, which combines all four. \Cref{fig:convergence_lemni} shows the running-average of tracking error (top) and the probability assigned to each expert by \AS (bottom). For the lemniscate target, the fast-adapting predictors \predA and \predB track best during the first 20$s$, while \predD, the worst predictor initially, becomes the clear best once its slower learning has converged. \AS captures this change where it first prefers \predA and \predB, then shifts its probability to \predD between 20-30$s$ to settle at about $0.85$ on \predD for the rest of the flight. As a result, the tracking error of our algorithm stays close to that of the best predictor at all times. 
The same ablation experiment with a randomized adversarial target is shown in \Cref{fig:convergence_rdm_adv}. In contrast with the lemniscate trajectory, \predD is now the worst predictor. \AS lowers its probability below $0.1$ within the first 30$s$ and mostly prefers \predB and \predA, the two best individual predictors. Interestingly, the bandit-learning combination of all four experts reaches a lower mean error ($0.333$\,m) than any individual predictor (best: \predB, $0.343$\,m), averaged across 10 runs. These results demonstrate that \AS converges to the best predictor(s) for the given target trajectory based on learning capacity and progress, and that it matches or improves on the best single predictor.

\section{Conclusion} \label{sec:con}

We presented a self-adaptive online learning for control method for tracking unknown and possibly adversarial target dynamics that can exhibit structured, random, and/or adversarial patterns. 
The method simultaneously learns multiple prediction models from scratch, via one-shot, self-supervised learning, and updates a probabilistic policy in real time to select the best one to match the observed target behavior. 
We provide theoretical no-regret guarantees to asymptotically match a clairvoyant optimal controller. Additionally, the proved guarantees degrade gracefully as a function of modeling error and switching target dynamics. 

\myParagraph{Future work} 
\textit{First}, we will extend our theoretical analysis to capture the performance guarantees to the case where the predictors' feature vectors include the robot state.
\textit{Second}, we will enhance the proposed approach to enable self-adaptation of the library of predictors, adding and learning new predictors upon detection of ``out-of-distribution" target behaviors.  
\textit{Third}, we will enable the method to choose predictors with risk-awareness, upon quantifying on-the-fly the uncertainty risk of each predictor.
\textit{Finally}, we will extend the framework to the decentralized case, enabling transfer learning while actively managing the adverse effects that communication delays can have on scalability and reliability.




\bibliographystyle{SageH}
\bibliography{references.bib}


\section{Appendix}\label{sec:app}

\subsection{Proof of Theorem \ref{theorem:regret}}\label{app:theorem_regret}

Using the performance difference lemma~\citep{kakade2003sample}, the regret can be equivalently written as 
\begin{equation}
    \begin{aligned}
        \DReg &\triangleq J_{T}\left(\pi; y_{1:T}\right)-J_{T}\left(\pi^\star; y_{1:T}\right) \\
        &= \sum_{t=1}^{T} \boldsymbol{A}_{t}^{\star}(u_t ; x_t).
    \end{aligned}
\end{equation}

By definition of $\boldsymbol{A}_{t}^{\star}(u_t ; x_t)$, we have
\begingroup
\allowdisplaybreaks
    \begin{align}
        \boldsymbol{A}_{t}^{\star}(u_t ; x_t) &= \left(\|e_t\|^2_{Q} + \|u_t\|^2_{R} + \boldsymbol{V}_{t+1}^{\star}(e_{t+1}) \right) \\
        &\qquad - \left(\|e_t\|^2_{Q} + \|u_t^{\star}\|^2_{R} + \boldsymbol{V}_{t+1}^{\star}(e_{t+1}^{\star}) \right) \\
        &= \|u_t\|^2_{R} - \|u_t^{\star}\|^2_{R} +  \boldsymbol{V}_{t+1}^{\star}(e_{t+1}) -  \boldsymbol{V}_{t+1}^{\star}(e_{t+1}^{\star}) \\
        &\leq L_{R} \| u_t - u_t^\star \| + L_V \| e_{t+1} - e_{t+1}^\star \| \\
        &= L_{R} \| u_t - u_t^\star \|  \\
        & \qquad + L_V \| x_{t+1} - y_{t+1} - x_{t+1}^\star + y_{t+1} \| \\
        &= L_{R} \| u_t - u_t^\star \| + L_V \| g(x_{t}) (u_t - u_t^\star) \| \\
        &\leq L_u \| u_t - u_t^\star \|
    \end{align}
\endgroup
where the first inequality holds by $u\in\calU$, where $\calU$ is a compact set and Lipschitzness of value function~(\Cref{assump:lipschitz}), the second inequality holds by boundedness of pursuer state~(\Cref{assump:bounded_state}) and locally Lipschitzness of $g(\cdot)$, and $L_{R}$ and $L_u$ are appropriate Lipschitz constant. Here, with a slight abuse of notation, $x_{t+1}^\star$ and $e_{t+1}^\star$ denote the state and the tracking error reached by applying $u_t^\star$, the control input of $\pi^\star$ at the current state $x_t$, \ie
\begin{equation}
    x_{t+1}^\star \triangleq f(x_t)+g(x_t)u_t^\star, \quad e_{t+1}^\star \triangleq x_{t+1}^\star - y_{t+1}.
\end{equation}

By \Cref{assump:perturbation}, we can relate $\| u_t - u_t^\star \|$ to the prediction errors of future target states:
\begin{equation}
    \begin{aligned}
        \boldsymbol{A}_{t}^{\star}(u_t ; x_t) &\leq L_u \| u_t - u_t^\star \| \leq L_u L_\pi \sum_{k=t+1}^{t+N} \|y_k -\hat{y}_k \| \\
        &= L_y \sum_{j=0}^{N-1} \|y_{t+1+j} -\hat{y}_{t+1+j} \|,
    \end{aligned}
\end{equation}
where $L_y \triangleq L_u L_\pi$ is independent of $t$ and $T$.

The above inequality shows that the advantage of $\pi^\star$ over $\pi$ at each time step is upper bounded by the prediction errors of $\hat{h}_j$, $j\in \{0, \dots, N-1\}$, with a constant scale $L_y$.

Then, the regret is upper bounded by
\begin{equation}
    \begin{aligned}
        \DReg &= \sum_{t=1}^{T} \boldsymbol{A}_{t}^{\star}(u_t ; x_t) \\
        &\leq L_y \sum_{t=1}^{T} \sum_{j=0}^{N-1} \|y_{t+1+j} -\hat{y}_{t+1+j} \| \\
        &\leq L_y  \sum_{t=1}^{T} \sqrt{N} \sqrt{ \sum_{j=0}^{N-1} \|y_{t+1+j} -\hat{y}_{t+1+j} \|^2},
    \end{aligned}
    \label{eq:regret_prediction}
\end{equation}
where the second inequality holds by the Cauchy–Schwarz inequality. 
To relate \cref{eq:regret_prediction} to the selection by \AS, for each predictor $p$ we define its prediction error and the corresponding prediction accuracy at time $t$,
\begin{equation}
    \begin{aligned}
        e_t^p &\triangleq \sqrt{ \sum_{j=0}^{N-1} \|y_{t+1+j} -\hat{y}_{t+1+j}^{p} \|^2}, \\
        r_t^p &\triangleq \bar{r} - e_t^p, \quad  \hat{y}_{t+1+j}^{p} \triangleq \hat{h}^p_j\left(y_{t-\tau:t}; \hat{\alpha}^{j,p}_{t}\right),
    \end{aligned}
    \label{eq:pred_error}
\end{equation}
where $\hat{y}_{t+1+j}^{p}$ is the prediction made by predictor $p$ at time $t$, and $r_t^p \in [0, \bar{r}]$. Without loss of generality, we let $\bar{r}=1$ (otherwise, we rescale $\lambda$ by $1/\bar{r}$). Since \MPC at time $t$ uses the predictor $p_t$ selected by \AS, \ie $p_t\sim\Delta_t$ is the predictor $\tilde{p}$ sampled in \Cref{alg:meta} at time $t$, \cref{eq:regret_prediction} becomes
\begin{equation}
    \DReg \leq L_y \sqrt{N} \sum_{t=1}^{T} e_t^{p_t}.
    \label{eq:regret_selected}
\end{equation}

The prediction accuracy $r_t^p$ differs from the one computed by \AS in \cref{eq:rmse_theory} in two aspects. First, $r_t^p$ depends on $y_{t+1:t+N}$, and thus becomes available only at time $t+N$, \ie  the \AS receives its feedback with a delay of $N$ time steps. Second, at time $t+N$, \cref{eq:rmse_theory} evaluates the predictions with $\hat{\alpha}^{j,p}_{t+N}$ instead of $\hat{\alpha}^{j,p}_{t}$. We denote the prediction error and accuracy computed by \AS at time $t+N$ as
\begin{equation}
    \begin{aligned}
        \tilde{e}_t^p &\triangleq \sqrt{ \sum_{j=0}^{N-1} \|y_{t+1+j} -\hat{h}^p_j\left(y_{t-\tau:t}; \hat{\alpha}^{j,p}_{t+N}\right) \|^2}, \\
        \tilde{r}_t^p &\triangleq \bar{r} - \tilde{e}_t^p,
    \end{aligned}
    \label{eq:pred_error_as}
\end{equation}
\ie $\tilde{r}_t^p$ is the reward $r_{t+N}^p$ used in \Cref{alg:meta} at time $t+N$. Since \cref{eq:rmse_theory} requires $N+\tau$ past target states, \Cref{alg:meta} receives no reward before time $N+1$, and thus $\Delta_t=\Delta_0$ for $t\leq N$.
By \Cref{assump:bounded_state} and the boundedness of the random Fourier features and of $\calD_j$, there exist constants $D$, $G$, and $B_\Phi$ such that, for all $\alpha,\alpha'\in\calD_j$,
\begin{equation}
    \begin{aligned}
       & \|\alpha-\alpha'\| \leq D, \qquad \|\nabla l_{t}^{j,p}(\alpha)\|\leq G, \\
        & \|\hat{h}^p_j(z;\alpha)-\hat{h}^p_j(z;\alpha')\| \leq B_\Phi\|\alpha-\alpha'\|,
    \end{aligned}
    \label{eq:constants}
\end{equation}
where the last inequality holds since $\hat{h}^p_j$ is linear in $\alpha$ per \cref{eq:kernal_approx}. We bound the difference between $\tilde{e}_t^p$ and $e_t^p$ as follows.

\begin{lemma}[Parameter Drift]\label{lemma:drift}
    For all $t$ and $k\geq 0$, $\|\hat{\alpha}^{j,p}_{t+k}-\hat{\alpha}^{j,p}_{t}\|\leq k\eta G$ and $|\tilde{e}_t^p - e_t^p|\leq N^{\frac{3}{2}}B_\Phi \eta G$.
\end{lemma}

We next utilize the performance guarantee of \Cref{alg:meta}, \ie the Fixed-Share algorithm~\citep{herbster1998tracking}, under the delayed rewards $\tilde{r}_t^p$.
\begin{proposition}[Regret Bound of Fixed-Share with Delayed Feedback]\label{theorem:hedge}
    Consider \Cref{alg:meta} with $\lambda>0$ and $\beta\in(0,1)$, where the reward $\tilde{r}_t^p\in[0,1]$ in \cref{eq:pred_error_as} is used at time $t+N$. Then, for any sequence of predictors $p^\star_1,\dots,p^\star_T$ that switches at most $\kappa$ times,
    \begin{equation}
        \begin{aligned}
            &\sum_{t=1}^{T} \tilde{r}_t^{p^\star_t} - \sum_{t=1}^{T} \sum_{p=1}^{P}\Delta_t^p\, \tilde{r}_t^{p} \\
            \leq\; & \frac{1}{\lambda}\left((\kappa+1)\ln P + \kappa \ln\frac{1}{\beta} + T\ln\frac{1}{1-\beta}\right) \\
            & + \frac{\lambda T}{8} + 2NT(\lambda+\beta).
        \end{aligned}
        \label{eq:fixed_share_regret}
    \end{equation}
    In particular, for $\lambda=\calO\left({1}/{\sqrt{T}}\right)$ and $\beta=\calO\left({1}/{T}\right)$, the right-hand side is $\calO\left(\left((\kappa+1)\ln P + \kappa\ln T + N\right)\sqrt{T}\right)$.
\end{proposition}

We now use \Cref{theorem:hedge} to replace the prediction error $e_t^{p_t}$ of the selected predictor $p_t$ in \cref{eq:regret_selected} with the prediction error $e_t^{p}$ of a fixed predictor $p$. Note that \Cref{theorem:hedge} controls the errors $\tilde{e}_t^p$ that \AS observes, whereas \cref{eq:regret_selected} involves the errors $e_t^p$ of the predictions used by \MPC. We therefore use \Cref{lemma:drift} to move from $e_t^p$ to $\tilde{e}_t^p$, apply \Cref{theorem:hedge}, and use \Cref{lemma:drift} again to move back to $e_t^p$.

By \cref{eq:target}, the target dynamics depend only on the history of the target states, and thus the target trajectory $y_{1:T}$ does not depend on the pursuer's actions or on the selections $p_{1:T}$. Hence, $y_{1:T}$ is fixed, and so are the parameters $\hat{\alpha}^{j,p}_t$ of \ISO (which depend only on $y_{1:T}$), the errors $e_t^p$ and $\tilde{e}_t^p$, and the policy $\Delta_t$ (which depends only on $\tilde{r}_t^p$). The only randomness is the selection $p_t\sim\Delta_t$, hence 
\begin{equation}
    \mathbb{E}\left[e_t^{p_t}\right]=\sum_{p=1}^{P}\Delta_t^p\, e_t^p.
    \label{eq:expectation_selected}
\end{equation}

Next, since $\bar{r}=1$, we have $\tilde{r}_t^q=1-\tilde{e}_t^q$, and thus $\sum_q\Delta_t^q\tilde{r}_t^q=1-\sum_q\Delta_t^q\tilde{e}_t^q$ since $\sum_q\Delta_t^q=1$. Hence, the regret in terms of rewards in \cref{eq:fixed_share_regret} equals the regret in terms of errors:
\begin{equation}
    \sum_{t=1}^{T} \tilde{r}_t^{p} - \sum_{t=1}^{T} \sum_{q=1}^{P}\Delta_t^q\, \tilde{r}_t^{q} = \sum_{t=1}^{T} \sum_{q=1}^{P}\Delta_t^q\, \tilde{e}_t^{q} - \sum_{t=1}^{T} \tilde{e}_t^{p}.
    \label{eq:reward_to_error}
\end{equation}
Applying \Cref{theorem:hedge} with $\kappa=0$, \ie $p^\star_t=p$ for all $t$, the right-hand side of \cref{eq:reward_to_error} is at most $\calO\left(\left(\ln P + N\right)\sqrt{T}\right)$. Together with \Cref{lemma:drift}, we have
\begin{equation}
    \begin{aligned}
        \mathbb{E}\left[\sum_{t=1}^{T} e_t^{p_t}\right] &\leq \sum_{t=1}^{T} \sum_{q=1}^{P}\Delta_t^q\, \tilde{e}_t^q + N^{\frac{3}{2}}B_\Phi \eta G T \\
        &\leq \sum_{t=1}^{T} \tilde{e}_t^{p} + \calO\left(\left(\ln P + N\right)\sqrt{T}\right) + N^{\frac{3}{2}}B_\Phi \eta G T\\
        &\leq \sum_{t=1}^{T} e_t^{p} + \calO\left(\left(\ln P + N + N^{\frac{3}{2}}B_\Phi G \right)\sqrt{T}\right),
    \end{aligned}
    \label{eq:fs_applied}
\end{equation}
where the first inequality holds by \cref{eq:expectation_selected} and \Cref{lemma:drift}, \ie $e_t^q\leq\tilde{e}_t^q+N^{\frac{3}{2}}B_\Phi\eta G$ for each $q$, averaged with the weights $\Delta_t^q$ that sum to one; the second inequality holds by \cref{eq:reward_to_error} and \Cref{theorem:hedge}; and the last inequality holds by \Cref{lemma:drift}, \ie $\tilde{e}_t^p\leq e_t^p+N^{\frac{3}{2}}B_\Phi\eta G$, and $N^{\frac{3}{2}}B_\Phi \eta G T=\calO(N^{\frac{3}{2}}B_\Phi G\sqrt{T})$ for $\eta=\calO\left({1}/{\sqrt{T}}\right)$. Combining \cref{eq:regret_selected} and \cref{eq:fs_applied}, we have
\begin{equation}
    \begin{aligned}
        \mathbb{E}\left[\DReg\right] \leq L_y  \sqrt{N}  \sum_{t=1}^{T} &\sqrt{ \sum_{j=0}^{N-1} \|y_{t+1+j}^{} -\hat{y}_{t+1+j}^{p} \|^2} \\ & \quad \quad \quad + \calO\left(\sqrt{T}\right).
    \end{aligned}
    \label{eq:reg_intermsof_reward}
\end{equation}

Since ${p} \in \{1, \dots, P\}$ can be chosen arbitrarily, let $p$ be the predictor that satisfies \Cref{assump:small_approx_error}. We then leverage regret bound of online least-squares estimation for the term $\sum_{t=1}^{T} \sqrt{ \sum_{j=0}^{N-1} \|y_{t+1+j}^{} -\hat{y}_{t+1+j}^{p} \|^2}$.

\begin{proposition}[Regret Bound of Online Least-Squares Estimation~\citep{hazan2016introduction}]\label{theorem:OGD}
    Assume $\eta=\calO\left({1}/{\sqrt{T}}\right)$ for predictor $p$.  Then,
    \begin{equation}
       \RegOGD\triangleq \sum_{t=1}^{T} l_t^{j,p} \left(\alpha_t^{j,p}\right) - \sum_{t=1}^{T} l_t^{j,p} \left(\alpha^{j,p}_{\star}\right)  \leq \calO\left(\sqrt{T}\right),
       \label{eq:prop2}
    \end{equation}
    where $\alpha_{\star}^{j,p} \triangleq \underset{\alpha \in \calD_{j}}{\operatorname{\textit{argmin}}}\;\sum_{t=1}^{T} l_t^{j,p} \left(\alpha\right)$ is the optimal parameter that achieves lowest cumulative loss in hindsight.
\end{proposition}

Note that $\hat{y}_{t+1+j}^{p}$ is computed with $\hat{\alpha}^{j,p}_{t}$, whereas the loss $l_{t+1+j}^{j,p}$ in \Cref{alg:expert}, whose feature vector is $y_{t-\tau:t}$ and target is $y_{t+1+j}$, is evaluated by \OGD at $\hat{\alpha}^{j,p}_{t+1+j}$. Since $l_{t+1+j}^{j,p}$ is $G$-Lipschitz on $\calD_j$, \Cref{lemma:drift} gives
\begin{equation}
    \begin{aligned}
        \|y_{t+1+j} -\hat{y}_{t+1+j}^{p} \|^2 &= l_{t+1+j}^{j,p}\left(\hat{\alpha}_t^{j,p}\right) \\
        &\leq l_{t+1+j}^{j,p}\left(\hat{\alpha}_{t+1+j}^{j,p}\right) + N\eta G^2.
    \end{aligned}
    \label{eq:delayed_ogd}
\end{equation}

Therefore, we have
    \begin{align}
        &\mathbb{E}\left[\DReg\right] \notag \\
        \leq& L_y  \sqrt{N}  \sum_{t=1}^{T} \sqrt{ \sum_{j=0}^{N-1} \|y_{t+1+j}^{} -\hat{y}_{t+1+j}^{p} \|^2} + \calO\left(\sqrt{T}\right) \notag \\
        \leq& L_y  \sqrt{NT}  \sqrt{ \sum_{t=1}^{T}  \sum_{j=0}^{N-1} \|y_{t+1+j}^{} -\hat{y}_{t+1+j}^{p} \|^2} + \calO\left(\sqrt{T}\right) \notag \\
        =& L_y  \sqrt{NT}  \sqrt{  \sum_{j=0}^{N-1} \sum_{t=1}^{T}  \|y_{t+1+j}^{} -\hat{y}_{t+1+j}^{p} \|^2} + \calO\left(\sqrt{T}\right)  \notag \\
        \leq& L_y \sqrt{NT} \sqrt{\sum_{j=0}^{N-1} \sum_{t=1}^{T}l_{t+1+j}^{j,p}\left(\hat{\alpha}_{t+1+j}^{j,p}\right) + N^2\eta G^2 T} + \calO\left(\sqrt{T}\right) \notag\\
        \leq& \calO \left(T^{\frac{3}{4}} \right) \label{eq:regret_no_error}
    \end{align}
where the second inequality holds by the Cauchy–Schwarz inequality, the third inequality holds by \cref{eq:delayed_ogd}, and the last inequality holds by \Cref{assump:small_approx_error} and \Cref{theorem:OGD}, \ie $\sum_{t=1}^{T} l_{t+1+j}^{j,p}\left(\hat{\alpha}_{t+1+j}^{j,p}\right)  \leq \calO\left(\sqrt{T}\right)$, applying \Cref{theorem:OGD} to the $T$ losses $l_{2+j}^{j,p},\dots,l_{T+1+j}^{j,p}$, and $N^2\eta G^2 T = \calO\left(\sqrt{T}\right)$ for $\eta=\calO\left({1}/{\sqrt{T}}\right)$.
\qed

\subsection{Proof of Corollary \ref{cor:regret_with_error}}\label{app:cor_regret_with_error}
In the presence of error $\epsilon_t^j$, \Cref{assump:small_approx_error} is no longer valid. 
The term $\sum_{t=1}^{T} l_t^{j,p} \left(\alpha^{j,p}_{\star}\right)$, in \cref{eq:prop2} (\Cref{theorem:OGD}) is non-zero, and instead, since $\alpha^{j,p}_{\star}$ in \cref{eq:prop2} minimizes the cumulative loss, it is upper bounded by the sum of errors in \cref{eq:approx_error}, \ie $\sum_{t=1}^{T}\|\epsilon_t^j\|^2$. Rearranging the terms in \cref{eq:prop2}, we have 
\begin{equation}
       \sum_{t=1}^{T} l_{t+1+j}^{j,p} \left(\hat{\alpha}_{t+1+j}^{j,p}\right) \leq  \calO\left(\sqrt{T}\right) + \sum_{t=1}^{T}\|\epsilon_t^j\|^2.
\end{equation}
Using \cref{eq:regret_no_error} and $\sqrt{a+b}\leq\sqrt{a}+\sqrt{b}$, we have
\begin{equation}
    \begin{aligned}
        &\mathbb{E}\left[\DReg\right] \\
        &\leq L_y \sqrt{NT} \sqrt{\calO\left(\sqrt{T}\right) + \sum_{j=0}^{N-1}\sum_{t=1}^{T}\|\epsilon_t^j\|^2} + \calO\left(\sqrt{T}\right) \\
        &\leq \calO\left(T^{\frac{3}{4}}\right) + L_y\sqrt{N}\sqrt{T\sum_{t=1}^{T}\sum_{j=0}^{N-1}\|\epsilon_t^j\|^2},
    \end{aligned}
\end{equation}
which is the equation for \Cref{cor:regret_with_error}.
\qed

\subsection{Proof of Corollary \ref{cor:switching}}\label{app:cor_switching}
In the case of switching dynamics, both the best predictor and its best parameters may change across the intervals $I_1,\dots,I_{\kappa+1}$. We first extend \Cref{theorem:OGD} to any time interval.
\begin{lemma}[Regret of \OGD over Intervals]\label{lemma:ogd_interval}
    For any interval $I$ of consecutive time steps with $|I|\leq T$ and any $\alpha\in\calD_j$,
    \begin{equation}
        \sum_{t\in I} l_t^{j,p}\left(\hat{\alpha}_t^{j,p}\right) - \sum_{t\in I} l_t^{j,p}\left(\alpha\right) \leq \frac{D^2}{2\eta} + \frac{\eta G^2 |I|}{2}.
    \end{equation}
    In particular, for $\eta=\calO\left({1}/{\sqrt{T}}\right)$, the right-hand side is at most $c_0\sqrt{T}$, where $c_0$ is a constant that depends only on $D$ and $G$.
\end{lemma}

The bound is the standard regret bound of OGD with a fixed step size~\citep{zinkevich2003online,hazan2016introduction}, applied to the losses in $I$ with any initial point $\hat{\alpha}$, since the analysis only requires $\|\hat{\alpha}-\alpha\|\leq D$.
Notably, \Cref{lemma:ogd_interval} holds with a step size $\eta$ that is independent of the interval, and thus of $\kappa$.

Consider the sequence of predictors $p^\star_t\triangleq p^\star_k$ for $t\in I_k$, which switches at most $\kappa$ times over $t\in\{1,\dots,T\}$. Following the proof of \Cref{theorem:regret}, but applying \Cref{theorem:hedge} with the sequence $p^\star_1,\dots,p^\star_T$ that switches $\kappa$ times instead of a fixed predictor, we have
\begin{equation}
    \mathbb{E}\left[\DReg\right] \leq L_y \sqrt{N} \sum_{k=1}^{\kappa+1}\sum_{t\in I_k} e_t^{p^\star_k} + \tilde{\calO}\left((\kappa+1)\sqrt{T}\right).
    \label{eq:cor2_first}
\end{equation}
Next, we bound the prediction errors on each interval $I_k$. By the Cauchy–Schwarz inequality,
\begin{equation}
    \sum_{t\in I_k} e_t^{p^\star_k} \leq \sqrt{|I_k|}\sqrt{\sum_{j=0}^{N-1}\sum_{t\in I_k}\|y_{t+1+j} -\hat{y}_{t+1+j}^{p^\star_k} \|^2}.
\end{equation}
For each $j$, by \cref{eq:delayed_ogd}, \Cref{lemma:ogd_interval} on the shifted interval $\{t+1+j: t\in I_k\}$ with $\alpha=\alpha^{j}_{\star,k}$, and $l_{t+1+j}^{j,p^\star_k}(\alpha^{j}_{\star,k})=\|\epsilon_t^j\|^2$ for $t\in I_k$, we have
\begin{equation}
    \sum_{t\in I_k}\|y_{t+1+j} -\hat{y}_{t+1+j}^{p^\star_k} \|^2 \leq c_0\sqrt{T} + \sum_{t\in I_k}\|\epsilon_t^j\|^2 + N\eta G^2|I_k|.
\end{equation}
Summing over $j$, letting $E_k\triangleq\sum_{j=0}^{N-1}\sum_{t\in I_k}\|\epsilon_t^j\|^2$, and using $\sqrt{a+b+c}\leq\sqrt{a}+\sqrt{b}+\sqrt{c}$, we obtain
\begingroup
\allowdisplaybreaks
\begin{align}
    &\sum_{k=1}^{\kappa+1}\sum_{t\in I_k} e_t^{p^\star_k} \notag \\
    \leq& \sum_{k=1}^{\kappa+1}\left(\sqrt{c_0 N}\,T^{\frac{1}{4}}\sqrt{|I_k|} + N G \sqrt{\eta}\,|I_k| + \sqrt{|I_k| E_k}\right) \notag \\
    \leq& \sqrt{c_0 N}\sqrt{\kappa+1}\,T^{\frac{3}{4}} + N G \sqrt{\eta}\, T + \sqrt{T\sum_{t=1}^{T}\sum_{j=0}^{N-1}\|\epsilon_t^j\|^2},
\end{align}
\endgroup
where the last inequality holds by $\sum_{k=1}^{\kappa+1}|I_k|=T$ and the Cauchy–Schwarz inequality, \ie $\sum_{k=1}^{\kappa+1}\sqrt{|I_k|}\leq\sqrt{(\kappa+1)T}$ and $\sum_{k=1}^{\kappa+1}\sqrt{|I_k|E_k}\leq\sqrt{T\sum_{k=1}^{\kappa+1}E_k}$. Substituting into \cref{eq:cor2_first}, with $N G \sqrt{\eta}\, T = \calO(T^{\frac{3}{4}})$ for $\eta=\calO\left({1}/{\sqrt{T}}\right)$, we obtain \Cref{cor:switching}.
\qed

The dependence on $\sqrt{\kappa+1}$, instead of $\kappa^{\frac{1}{4}}$, is due to the step size $\eta=\calO\left(1/\sqrt{T}\right)$ that does not depend on $\kappa$, under which the regret of \OGD on each interval is $\calO(\sqrt{T})$ rather than $\calO(\sqrt{|I_k|})$ (\Cref{lemma:ogd_interval}). If an upper bound $\bar{\kappa}\geq\kappa$ is known, choosing $\eta=\calO\left(\sqrt{\bar{\kappa}/T}\right)$ and using the dynamic regret bound of \OGD~\citep{zinkevich2003online} improves the first term in \cref{eq:corollary_switching} to $\calO(\bar{\kappa}^{\frac{1}{4}}T^{\frac{3}{4}})$. We choose $\eta=\calO\left({1}/{\sqrt{T}}\right)$ so that the proposed method requires no knowledge of $\kappa$.

\subsection{Proof of \Cref{lemma:drift}}

For the first claim, each \OGD step in \Cref{alg:expert} moves the parameters by $\eta\nabla_t^j$, whose norm is at most $\eta G$ by \cref{eq:constants}, and the projection onto the convex set $\calD_j$ is non-expansive. Hence, $\|\hat{\alpha}^{j,p}_{s+1}-\hat{\alpha}^{j,p}_{s}\|\leq\eta G$ for all $s$, and the first claim follows by the triangle inequality over the $k$ steps from $t$ to $t+k$.

For the second claim, we stack the $N$ prediction errors into vectors in $\mathbb{R}^{N d_y}$, \ie $a\triangleq\left[a_0^\top,\dots,a_{N-1}^\top\right]^\top$ and $b\triangleq\left[b_0^\top,\dots,b_{N-1}^\top\right]^\top$ with $a_j\triangleq y_{t+1+j}-\hat{h}^p_j(y_{t-\tau:t};\hat{\alpha}^{j,p}_{t})$ and $b_j\triangleq y_{t+1+j}-\hat{h}^p_j(y_{t-\tau:t};\hat{\alpha}^{j,p}_{t+N})$, so that $e_t^p=\|a\|$ and $\tilde{e}_t^p=\|b\|$ by \cref{eq:pred_error} and \cref{eq:pred_error_as}. Then,
\begin{equation}
	\begin{aligned}
		|\tilde{e}_t^p - e_t^p| &= \big|\|b\|-\|a\|\big| \leq \|b-a\| \\
		&= \sqrt{\sum_{j=0}^{N-1} \left\|\hat{h}^p_j(y_{t-\tau:t};\hat{\alpha}^{j,p}_{t+N})-\hat{h}^p_j(y_{t-\tau:t};\hat{\alpha}^{j,p}_{t})\right\|^2} \\
		&\leq \sqrt{\sum_{j=0}^{N-1} B_\Phi^2 \left\|\hat{\alpha}^{j,p}_{t+N}-\hat{\alpha}^{j,p}_{t}\right\|^2} \\
		&\leq \sqrt{N B_\Phi^2 N^2\eta^2G^2} = N^{\frac{3}{2}}B_\Phi \eta G,
	\end{aligned}
	\label{eq:drift_proof}
\end{equation}
where the first inequality is the reverse triangle inequality, the second equality holds since the target states $y_{t+1+j}$ cancel in $b_j-a_j$, \ie the difference between the two prediction errors depends only on the difference between the two predictions, the second inequality holds by \cref{eq:constants}, since $\hat{h}^p_j$ is linear in the parameters, and the last inequality holds by the first claim with $k=N$, applied to each of the $N$ terms. \qed

\subsection{Proof of \Cref{theorem:hedge}}

We follow the analysis of classic Fixed-Share~\citep{herbster1998tracking}, \cite[Chapter~5]{cesa2006prediction}, under delayed feedback~\citep{joulani2016delay,korotin2020adaptive}. Since $\tilde{r}_t^p$ is used at time $t+N$, the updates in \cref{eq:hedge_update} and \cref{eq:fixed_share} read
\begin{equation}
	\begin{aligned}
		\bar{\Delta}_{t+N}^{p} &= \frac{\Delta_{t+N-1}^{p}\exp(\lambda \tilde{r}_{t}^{p})}{\sum_{q} \Delta_{t+N-1}^{q}\exp(\lambda \tilde{r}_{t}^{q})}, \\
		\Delta_{t+N}^{p} &= (1-\beta)\,\bar{\Delta}_{t+N}^{p} + \frac{\beta}{P}.
	\end{aligned}
	\label{eq:fs_shifted}
\end{equation}
That is, $\Delta_{t+N-1}$ is the last policy before the reward $\tilde{r}_t$ is used, and thus plays the role of the policy without delay. We first bound the regret of $\Delta_{t+N-1}$ (Steps~1--3), and then account for the delay, \ie that the predictor at time $t$ is sampled from $\Delta_t$ instead of $\Delta_{t+N-1}$ (Step~4). Throughout, we use $\ln(1/\Delta_s^p)$ as a potential function: it is small when the policy $\Delta_s$ concentrates on the predictor $p$, and large otherwise.

\textit{Step 1 (exponential-weights update).} Taking the logarithm of the first line of \cref{eq:fs_shifted}, for any predictor $p$,
\begin{equation}
	\begin{aligned}
		\ln\frac{\bar{\Delta}_{t+N}^{p}}{\Delta_{t+N-1}^{p}} &= \lambda \tilde{r}_{t}^{p} - \ln \sum_{q} \Delta_{t+N-1}^{q}\exp(\lambda \tilde{r}_{t}^{q}) \\
		&\geq \lambda \tilde{r}_{t}^{p} - \lambda \sum_{q} \Delta_{t+N-1}^{q}\tilde{r}_{t}^{q} - \frac{\lambda^2}{8},
	\end{aligned}
	\label{eq:fs_step1}
\end{equation}
where the inequality holds by Hoeffding's lemma, \ie $\ln\mathbb{E}[e^{\lambda X}]\leq\lambda\mathbb{E}[X]+\lambda^2/8$ for any random variable $X\in[0,1]$, applied to $X=\tilde{r}_t^{q}$ with $q\sim\Delta_{t+N-1}$. \Cref{eq:fs_step1} states that the weight of the predictor $p$ increases when its reward $\tilde{r}_t^p$ is higher than the average reward of the policy.

\textit{Step 2 (mixing).} By the second line of \cref{eq:fs_shifted}, since both terms are nonnegative,
\begin{equation}
	\Delta_{t+N}^{p} \geq (1-\beta)\,\bar{\Delta}_{t+N}^{p}, \qquad \Delta_{t+N}^{p} \geq \frac{\beta}{P}.
	\label{eq:fs_step2}
\end{equation}
Taking the logarithm of the first inequality in \cref{eq:fs_step2} gives $\ln(1/\bar{\Delta}_{t+N}^{p})\geq\ln(1/\Delta_{t+N}^{p})-\ln(1/(1-\beta))$, \ie mixing decreases the weight of the predictor $p$ by at most a factor $1-\beta$. Rewriting the left-hand side of \cref{eq:fs_step1} as $\ln(1/\Delta_{t+N-1}^{p})-\ln(1/\bar{\Delta}_{t+N}^{p})$ and using this inequality, we have
\begin{equation}
	\begin{aligned}
		&\lambda\Big(\tilde{r}_{t}^{p} - \sum_{q} \Delta_{t+N-1}^{q}\tilde{r}_{t}^{q}\Big) - \frac{\lambda^2}{8} \\
		\leq\; & \ln\frac{1}{\Delta_{t+N-1}^{p}} - \ln\frac{1}{\Delta_{t+N}^{p}} + \ln\frac{1}{1-\beta}.
	\end{aligned}
\end{equation}
Dividing by $\lambda$, we obtain, for any predictor $p$,
\begin{equation}
	\begin{aligned}
		\tilde{r}_{t}^{p} - \sum_{q} \Delta_{t+N-1}^{q}\tilde{r}_{t}^{q} \leq\; & \frac{1}{\lambda}\left(\ln\frac{1}{\Delta_{t+N-1}^{p}} - \ln\frac{1}{\Delta_{t+N}^{p}}\right) \\
		&+ \frac{1}{\lambda}\ln\frac{1}{1-\beta} + \frac{\lambda}{8}.
	\end{aligned}
	\label{eq:fs_step2_combined}
\end{equation}
The left-hand side of \cref{eq:fs_step2_combined} is the regret at time $t$ with respect to the predictor $p$. It is upper bounded by three terms: the decrease of the potential $\ln(1/\Delta^p)$, which telescopes over time (Step~3); the cost of mixing, which is small for small $\beta$; and the cost of sampling from Hoeffding's lemma, which is small for small $\lambda$. We write the right-hand side in terms of the mixed policy $\Delta_{t+N}$, rather than $\bar{\Delta}_{t+N}$, so that the potential telescopes and the lower bound $\beta/P$ in \cref{eq:fs_step2} can be used.

\textit{Step 3 (summing over time).} Setting $p=p^\star_t$ in \cref{eq:fs_step2_combined} and summing over $t$, the first term on the right-hand side becomes
\begin{equation}
	\begin{aligned}
		&\sum_{t=1}^{T}\left(\ln\frac{1}{\Delta_{t+N-1}^{p^\star_t}} - \ln\frac{1}{\Delta_{t+N}^{p^\star_t}}\right) \\
		=\; & \ln\frac{1}{\Delta_{N}^{p^\star_1}} - \ln\frac{1}{\Delta_{T+N}^{p^\star_T}} + \sum_{t=1}^{T-1}\left(\ln\frac{1}{\Delta_{t+N}^{p^\star_{t+1}}} - \ln\frac{1}{\Delta_{t+N}^{p^\star_t}}\right) \\
		\leq\; & \ln P + \kappa \ln\frac{P}{\beta},
	\end{aligned}
	\label{eq:fs_step3}
\end{equation}
where the equality regroups the sum so that the term $-\ln(1/\Delta_{t+N}^{p^\star_t})$ at time $t$ is paired with the term $\ln(1/\Delta_{t+N}^{p^\star_{t+1}})$ at time $t+1$. The inequality holds since (i) $\Delta_{N}^{p^\star_1}=\Delta_{0}^{p^\star_1}=1/P$, as no reward is used before time $N+1$; (ii) $\ln(1/\Delta_{T+N}^{p^\star_T})\geq 0$; and (iii) each summand of the last sum is zero if $p^\star_{t+1}=p^\star_t$, \ie if the comparator does not switch, and it is at most $\ln(1/\Delta_{t+N}^{p^\star_{t+1}})\leq\ln(P/\beta)$ otherwise, by the second inequality in \cref{eq:fs_step2}. Since there are at most $\kappa$ switches, the last sum is at most $\kappa\ln(P/\beta)$. Therefore, summing \cref{eq:fs_step2_combined} over $t$ with $p=p^\star_t$ and using \cref{eq:fs_step3} and $\ln(P/\beta)=\ln P+\ln(1/\beta)$,
\begin{equation}
	\begin{aligned}
		&\sum_{t=1}^{T} \tilde{r}_t^{p^\star_t} - \sum_{t=1}^{T} \sum_{q=1}^{P}\Delta_{t+N-1}^q\, \tilde{r}_t^{q} \\
		\leq\; & \frac{1}{\lambda}\left((\kappa+1)\ln P + \kappa \ln\frac{1}{\beta} + T\ln\frac{1}{1-\beta}\right) + \frac{\lambda T}{8}.
	\end{aligned}
	\label{eq:fs_step3_result}
\end{equation}

\textit{Step 4 (delay).} \Cref{eq:fs_step3_result} evaluates the policy $\Delta_{t+N-1}$, while the predictor at time $t$ is sampled from $\Delta_{t}$, which is $N-1$ updates older. We show that the policy changes slowly, so that this delay only incurs a small bounded cost. Each update from $\Delta_{s-1}$ to $\Delta_{s}$ consists of the exponential-weights update from $\Delta_{s-1}$ to $\bar{\Delta}_{s}$, and the mixing from $\bar{\Delta}_{s}$ to $\Delta_{s}$. 
For the exponential-weights update, by the first line of \cref{eq:fs_shifted}, $\bar{\Delta}_{s}^p = \Delta_{s-1}^p e^{\lambda\tilde{r}^p}/Z$, where $Z\triangleq\sum_q\Delta_{s-1}^q e^{\lambda\tilde{r}^q}$ and $\tilde{r}$ denotes the reward used at time $s$. Since $\tilde{r}^p\in[0,1]$, both $e^{\lambda\tilde{r}^p}$ and its weighted average $Z$ lie in $[1,e^{\lambda}]$, hence $e^{\lambda\tilde{r}^p}/Z\geq e^{-\lambda}$, and $\Delta_{s-1}^p-\bar{\Delta}_{s}^p\leq\left(1-e^{-\lambda}\right)\Delta_{s-1}^p$ for every predictor $p$. Moreover, since $\Delta_{s-1}$ and $\bar{\Delta}_{s}$ are both probability distributions, $\sum_p\left(\bar{\Delta}_{s}^p-\Delta_{s-1}^p\right)=0$, \ie the total increase of the weights equals their total decrease. Hence, we have
\begin{equation}
	\begin{aligned}
		\sum_{p=1}^{P}\left|\bar{\Delta}_{s}^p-\Delta_{s-1}^p\right| &= 2\sum_{p=1}^{P}\max\left\{\Delta_{s-1}^p-\bar{\Delta}_{s}^p,\,0\right\} \\
		&\leq 2\left(1-e^{-\lambda}\right)\sum_{p=1}^{P}\Delta_{s-1}^p = 2\left(1-e^{-\lambda}\right) \\
        & \leq 2\lambda,
	\end{aligned}
\end{equation}
where the last inequality holds since $e^{-\lambda}\geq 1-\lambda$ for all $\lambda\geq 0$, by the convexity of $e^{-\lambda}$. 
For the mixing, by the second line of \cref{eq:fs_shifted}, $\Delta_{s}^p-\bar{\Delta}_{s}^p=\beta(1/P-\bar{\Delta}_{s}^p)$, hence
\begin{equation}
	\sum_{p=1}^{P}\left|\Delta_{s}^p-\bar{\Delta}_{s}^p\right| \leq \beta\sum_{p=1}^{P}\left(\bar{\Delta}_{s}^p+\frac{1}{P}\right) = 2\beta.
\end{equation}
By the triangle inequality, for any $s$,
\begin{equation}
	\sum_{p=1}^{P}\left|\Delta_{s}^p-\Delta_{s-1}^p\right| \leq 2\left(1-e^{-\lambda}\right) + 2\beta \leq 2(\lambda+\beta).
\end{equation}
For $s\leq N$, no reward is used and $\Delta_s=\Delta_{s-1}$, so the bound holds trivially. Since $\tilde{r}_t^p\in[0,1]$, the difference between the rewards of the two policies is at most their $\ell_1$ distance, \ie
\begin{equation}
	\begin{aligned}
		\sum_{p=1}^{P}\left(\Delta_{t+N-1}^p-\Delta_{t}^p\right)\tilde{r}_t^p &\leq \sum_{s=t+1}^{t+N-1}\sum_{p=1}^{P}\left|\Delta_{s}^p-\Delta_{s-1}^p\right| \\
		&\leq 2N(\lambda+\beta).
	\end{aligned}
	\label{eq:fs_step4}
\end{equation}
Summing \cref{eq:fs_step4} over $t$ and adding it to \cref{eq:fs_step3_result} gives \cref{eq:fixed_share_regret}. Finally, for $\lambda=\calO\left({1}/{\sqrt{T}}\right)$ and $\beta=\calO\left({1}/{T}\right)$, we have $\kappa\ln(1/\beta)=\calO(\kappa\ln T)$, $T\ln\frac{1}{1-\beta}\leq \frac{T\beta}{1-\beta}=\calO(1)$, $\lambda T/8 = \calO(\sqrt{T})$, and $2NT(\lambda+\beta)=\calO(N\sqrt{T})$, which, together with $1/\lambda=\calO(\sqrt{T})$, gives the particular case.  \qed

\begin{figure*}[t]
    \centering
    \includegraphics[width=\linewidth]{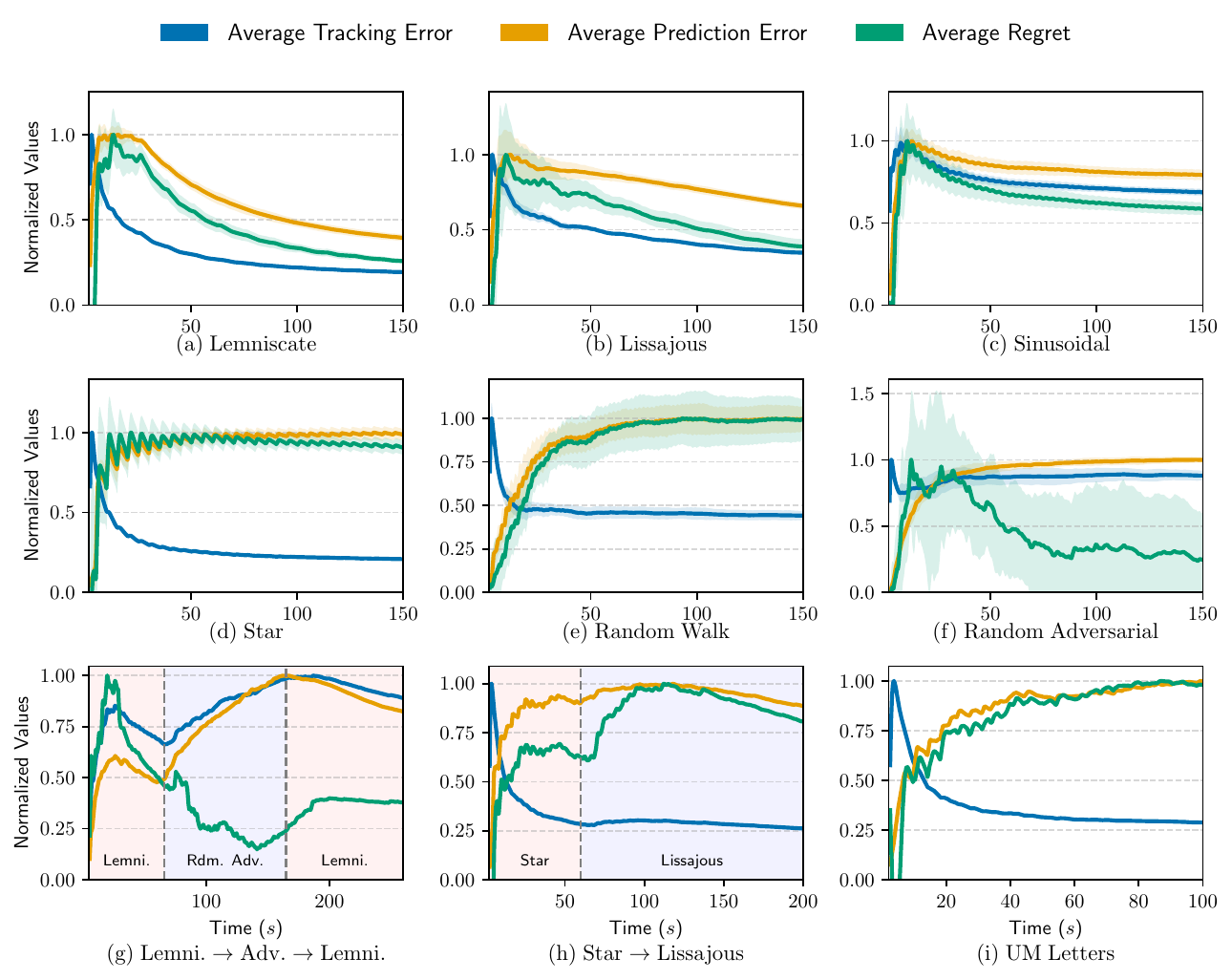}
    \caption{\textbf{Trends for regret, tracking error, and prediction error.}
    The normalized values for average tracking error, average prediction error, and average regret ($\text{Regret}_T^D/T$) for our algorithm across all target types considered in Section~\ref{subsec:sim_results} and Section~\ref{subsec:hardware}. The subplots~(a)-(f) are from simulation experiments in Section~\ref{subsec:sim_results} and the subplots~(g)-(i) are from hardware experiments in Section~\ref{subsec:hardware}.}
    \label{fig:regret_plot}
\end{figure*}

\subsection{Trends for Regret and Prediction Error}\label{app:regret}

In \Cref{fig:regret_plot}, we report the trends (normalized by their peaks) for average tracking error, average prediction error, and average regret of our algorithm against \OPT across all target types considered in \Cref{subsec:sim_results} and \Cref{subsec:hardware}.
In the proof of \Cref{theorem:regret}, we relate the regret to the prediction error in \cref{eq:regret_prediction}; therefore, we empirically examine their relationship here. As shown in \Cref{fig:regret_plot}, the average regret follows similar trends for all non-adversarial targets, which supports the theoretical results.

For the smooth trajectories, \ie (a) lemniscate, (b) Lissajous, and (c) sinusoidal, both the average regret and the average prediction error decrease over time, indicating a sublinear regret, per \Cref{theorem:regret}. The hardware experiment in (h) shows the same behavior: after the target switches from the star to the Lissajous trajectory at 60$s$, both quantities first rise and then decrease once the predictor has learnt and adapted to the new trajectory. For the switching and random trajectories, \ie the star (d), the random walk (e), and the UM letters (i), the average regret increases and levels off at a nonzero value, per \Cref{cor:switching}.

In the adversarial target scenarios, we observe the opposite: the average prediction error increases, while the average regret decreases. Unlike the other targets, the adversarial target reacts to the pursuer. Our theoretical framework does not cover adversarial trajectories, where the future motion depends on the pursuer state.
\OPT achieves better tracking, but spends considerably more control effort than our method, chasing the evading target whose speed is inversely determined by the distance to the pursuer, highlighted in adversarial policy from \ref{subsec:sim_results}. 
Therefore, the control cost of \OPT becomes significantly larger than ours and it causes the decrease of regret.

The average regret and the average tracking error do not necessarily align in \Cref{fig:regret_plot},
as regret is measured relative to \OPT and also includes the control cost of the \MPC cost function. Therefore, the regret guarantee in \Cref{theorem:regret} ensures the convergence of a cost that includes both tracking error and control effort, measured relative to \OPT.

\subsection{Implementation Details}\label{app:impl_details}
\myParagraph{MPC Setup} 
We use a nonlinear tracking \MPC with $N=10$ steps for a future time horizon of 1$s$. The control frequency is fixed at $50$ Hz. The quadratic cost weights for position and velocity respectively are $Q_p=20I$ and $Q_v=\text{diag}\left(5,5,25\right)$. Similarly, the quadratic control cost is set as $R=\text{diag}\left(100,100,100,100\right)$. The target motion model is first used to predict the target velocity and the target trajectory is obtained via Euler integration, which is set as a reference for the tracking \MPC. We adopt the \MPC framework from \cite{llanes2024crazysim}, which uses \textit{acados}~\citep{verschueren2021} and \textit{CasADi}~\citep{andersson2019}. 
We use RK4 to obtain the discretized dynamics. 

\myParagraph{Software and Hardware Setup} 
The software-in-the-loop (SITL) setup for the Crazyflie, based on \cite{llanes2024crazysim}, uses Gazebo/ROS2 to simulate target and pursuer drones. The pursuer can only access the past and current positions of the target and use the positions to predict its future motion. 
For the hardware experiments, we use two physical Crazyflie drones and the Lighthouse Positioning System for localization.
Simulations and computations are carried out on a laptop with an Intel i9-13900H CPU and 32GB RAM.

\myParagraph{Benchmark Algorithms}
\PLOT predicts target positions with linear motion models $y_{t+j} = C_j y_{t-\tau:t}$, where $C_j$ is updated online with recursive least-squares, that requires tuning the forgetting factor $\gamma \in [0, 1]$. After testing multiple values, we report the results only for $\gamma=0.5$ and $\gamma=0.9$. For a particular target policy, we observe that either small or large $\gamma$ provides best performance.
\SSI uses the random Fourier features for predictions of target motion, which forms a subroutine inside the \ISO module for our algorithm. To ensure fair comparison, we choose $M_{\SSI} = M =50$ sample points for the \SSI algorithm. We report results for values of std. dev. of Gaussian kernel $\sigma=0.1$ and $\sigma=1.0$ as they represent the spectrum of best performance for the tested scenarios.
\MODL uses three learners -- online linear regression, a multi-layer perceptron (MLP) and a set-learning residual neural network in tandem. We use a 4-layer MLP with a constant width of 128. For the set-learner, we use the proposed architecture with 2 blocks, each with 2 layers and width of 64 neurons. This choice facilitates real-time learning in our scenario, within the time-scale for our presented results.

\myParagraph{Algorithm Parameters}
For our algorithm, we select the kernel $K$ as a Gaussian kernel. To approximate the prediction function $\hat{h}\left(\cdot\right)$, we choose $M=50$ sample points. These $M$ random features $\left\{\theta_i\right\}_{i=1}^M$ are obtained i.i.d. from sampling $w_i$ from a Gaussian distribution $\mathcal{N}\left(0,\sigma^2\right)$, and $b_i$ from a uniform distribution of $\left[0,2\pi\right)$~\citep{rahimi2007random}. We use cosine function for $\phi: \mathbb{R} \rightarrow[-1,1]$. The standard deviation $\sigma$ is an important parameter that significantly affects the prediction and tracking performance. The learning rate $\eta$ for the \OGD step (\Cref{alg:expert}), and the number of steps $\tau$ for the target state history in the feature space also play an important role in the prediction performance of the single predictor, as we discuss in \ref{subsec:ablation_study}. 
We vary these parameters to obtain four ($P=4$) predictors that respond differently to different target behaviors, as we discuss in this section.
All four predictors use the same memory horizon $\tau=10$ and differ in their learning rate $\eta$ and kernel width $\sigma$:
\scenario{P1}: $\{\eta=0.4,\; \sigma=0.03\}$,
\scenario{P2}: $\{\eta=0.2,\; \sigma=0.1\}$,
\scenario{P3}: $\{\eta=0.1,\; \sigma=0.3\}$ and
\scenario{P4}: $\{\eta=0.05,\; \sigma=1.0\}$.
The predictors range from fast-adapting with a narrow kernel (\scenario{P1}), which tracks abrupt changes in the target's motion, to slow-adapting with a wide kernel (\scenario{P4}), which gives smoother and more accurate predictions for regular, repeating motion.

For the \ISO module,
in each of the \textit{P} independent predictors, the learned target dynamics over the horizon, are characterized by functions $\hat{h}^p_j\left(\cdot~; \hat{\alpha}^{j,p}_{t}\right)\in \mathbb{R}^3$, where $p\in\{1, \dots, P\}$ and $j\in\{0,N-1\}$. Each of these functions predicts the target velocity for \textit{j}-th step, which is then used to predict target position $\hat{y}_{t+j+1}\in\mathbb{R}^3$. 
Specifically, $\hat{v}_{k}^{p} = \frac{\hat{y}_k^{p} - \hat{y}_{k-1}^{p}}{\delta t}$, $v_{k} = \frac{y_k - y_{k-1}}{\delta t}$ represent the predicted target velocity and ground-truth target velocity respectively, and $\delta t$ is the \MPC sampling time.

For the \AS module, the prediction accuracy is defined as the error in the target velocity, since each $\hat{h}^p_j\left(\cdot~; \hat{\alpha}^{j,p}_{t}\right)$ predicts the target velocity. 
In \Cref{alg:meta}, we choose learning rate $\lambda=2.0$, and share rate $\beta=0.05$. 

\myParagraph{Choice of hyperparameters}  We provide intuition and guidance on choosing the hyperparameters for our algorithm. 
\begin{itemize}
    \item \myParagraph{Number of predictors ($P$)} The user may choose as many predictors as the computational platform allows for real-time execution. Additional predictors do not have a negative impact apart from the initial exploration cost for the \AS module. The computational cost generally scales linearly with $P$. For different number of predictors ($M=50$ and $\tau=10$), the run-time for each iteration of \ISO and \AS was found to be $2.78\pm1.15\;ms$ ($P=1$), $4.97\pm1.98\;ms$ ($P=2$) and $9.02\pm3.24\;ms$ ($P=4$), allowing it to run during $50Hz$ MPC loop. 
    \item \myParagraph{History window ($\tau$)} This affects the feature size and past information context during the inference. For example, a lower value such as $\tau=2$ would be insufficient for predicting motion for a lemniscate-like trajectory that involves overlapping points. In contrast, having a large history window can be disadvantageous in the case of highly unstructured or adversarial motion. As enabled by our framework, a user can choose predictors with different values for $\tau$. In our case, the runtime for each iteration of prediction was $2.06\pm0.97\;ms$ ($\tau=1$), $2.21\pm1.04\;ms$ ($\tau=5$), and $2.78\pm1.15\;ms$ ($\tau=10$).
    \item \myParagraph{Number of random Fourier features ($M$)} This represents the finite-dimensional approximation of the dynamics model $h(\cdot)$. As such, higher values of $M$ will always lead to accurate models, but poses as the main computational bottleneck for the algorithm. For $\tau=10$ and $P=1$, we found the per-iteration run-time to be $2.3\pm1.05\;ms$ ($M=25$), $2.78\pm1.85\;ms$ ($M=50$), $16.56\pm4.13\;ms$ ($M=65$) and $25.23\pm7.89\;ms$ ($M=75$). 
    \item \myParagraph{Learning rate for \OGD ($\eta$)} Similar to other learning algorithms, a lower value leads to slow learning while high values lead to unstable behavior. We found the values ranging from $\eta\in[0.05,0.25]$ to have good tracking and prediction performance. The user can select values to design a fast learner that adapts quickly to switching target dynamics.
    \item \myParagraph{Std. dev. for Gaussian kernel ($\sigma$)} Low values of $\sigma$ leads to narrow kernels, which results in learner with high variance and low bias. This can be beneficial in case of highly non-stationary target motion. In contrast, noisy trajectories show better performance with high values of $\sigma$. We tested our method with values $\sigma\in[0.075,1.5]$, showing good performance depending on the target type.
    \item \myParagraph{Share rate for Fixed-Share ($\beta$)} This relates to the classic exploration-exploitation tradeoff in learning algorithms. A larger $\beta$ keeps a larger minimum selection probability $\beta/P$ for each predictor, which enables faster adaptation to switching target dynamics at the cost of more exploration. The user may select $\beta$ based on the expected time-scale of switching dynamics. In our experiment, we use $\beta=0.05$.
    \item \myParagraph{Learning rate for \FSF ($\lambda$)} This value depends on the frequency of the \FSF updates and specific application. In our case, where we update the \FSF policy every $0.25\;s$, we found $\lambda\in[1.0,4.0]$ showing good performance.
    \end{itemize}

\end{document}